\documentclass{article}

\usepackage{arxiv}

\usepackage[utf8]{inputenc} 
\usepackage[T1]{fontenc}    
\usepackage{hyperref}       
\usepackage{url}            
\usepackage{booktabs}       
\usepackage{amsfonts}       
\usepackage{nicefrac}       
\usepackage{microtype}      
\usepackage{lipsum}         
\usepackage{graphicx}
\usepackage{natbib}
\usepackage{doi}
\usepackage{multirow}
\usepackage{multicol}
\usepackage{amsmath}
\usepackage{amssymb}

\usepackage{cleveref}       

\title{Trustworthy Representation Learning Across Domains}

\newif\ifuniqueAffiliation
\uniqueAffiliationtrue

\ifuniqueAffiliation 
\author{
        Ronghang Zhu \\
        University of Georgia \\
        \texttt{ronghangzhu@uga.edu} \\
        \And
        Dongliang Guo \\
        University of Virginia \\
        \texttt{dongliang.guo@virginia.edu} \\
        \And
        Daiqing Qi \\
        University of Virginia \\
        \texttt{daiqingqi@virginia.edu} \\
        \And
        Zhixuan Chu \\
        Ant Group \\
        \texttt{chuzhixuan.czx@alibaba-inc.com} \\
        \And
        Xiang Yu \\
        Amazon, Prime Video \\
        \texttt{yuxiang03@gmail.com} \\
        \And
        Sheng Li \\
        University of Virginia \\
        \texttt{shengli@virginia.edu} \\
}
\else
\usepackage{authblk}

\newbox{\orcid}\sbox{\orcid}{\includegraphics[scale=0.06]{orcid.pdf}} 
\fi

\renewcommand{\shorttitle}{\textit{arXiv} Template}

\hypersetup{
pdftitle={A template for the arxiv style},
pdfsubject={q-bio.NC, q-bio.QM},
pdfauthor={Ronghang Zhu,Dongliang Guo, Daiqing Qi, Zhixuan Chu, Xiang Yu, Sheng Li},
pdfkeywords={First keyword, Second keyword, More},
}

\begin{document}
\maketitle

\begin{abstract}
As AI systems have obtained significant performance to be deployed widely in our daily live and human society, people both enjoy the benefits brought by these technologies and suffer many social issues induced by these systems. To make AI systems good enough and trustworthy, plenty of researches have been done to build guidelines for trustworthy AI systems. Machine learning is one of the most important parts for AI systems and representation learning is the fundamental technology in machine learning. How to make the representation learning trustworthy in real-world application, e.g., cross domain scenarios, is very valuable and necessary for both machine learning and AI system fields. Inspired by the concepts in trustworthy AI, we proposed the first trustworthy representation learning across domains framework which includes four concepts, i.e, robustness, privacy, fairness, and explainability, to give a comprehensive literature review on this research direction. Specifically, we first introduce the details of the proposed trustworthy framework for representation learning across domains. Second, we provide basic notions and comprehensively summarize existing methods for the trustworthy framework from four concepts. Finally, we conclude this survey with insights and discussions on future research directions.
\end{abstract}


\section{Introduction}

Representation learning stands as a cornerstone challenge within the realm of machine learning. Its primary purpose is to encapsulate and delineate data in a manner conducive to the application of machine learning techniques. The efficacy of machine learning algorithms is profoundly contingent upon the caliber of data representation. Invariably, the potency of these algorithms is closely intertwined with the discriminative nature of the representations learned. Notably, representations endowed with heightened discriminative attributes consistently yield superior performance in contrast to their less-discriminative counterparts. As a consequence, much of the efforts have been devoted to feature engineering works which aim to design a good feature representation from raw data that can benefit machine learning. Early feature engineering works are highly depended on domain specific and human ingenuity. For example, in computer vision, some hand-designed feature descriptors, e.g., Local Binary Pattern (LBP)~\citep{TIP10-LBP}, Scale Invariant Feature Transform (SIFT)~\citep{ICCV99-SIFT}, and Histogram of Oriented Gradients (HOG)~\citep{CVPR05-HoG}, are proposed to extract the representations of the data for recognition and detection related tasks. However, these well-designed descriptors are heavily relied on human effort and domain specialist knowledge. In order to reduce the dependence on feature engineering, an automatic representation learning strategy, i.e., deep learning, is proposed to reduce the need of human labor and the requirement of specific expert knowledge. Deep learning~\citep{ACM18-DL-Survey,NeurComp16-DL-Survey} has emerged as an omnipresent facet of representation learning, finding extensive application across a plethora of real-world domains including but not limited to image classification, video captioning, and social network analysis. The confluence of burgeoning data availability and remarkable advancements in high-performance computing resources has catalyzed the proliferation of deep learning frameworks. Noteworthy examples encompass Convolutional Neural Network (CNN)\cite{CNN-LeCun}, Recurrent Neural Network (RNN)\cite{RNN_MJ}, Graph Neural Network (GNN)\cite{TNN08-GNN}, Generative Adversarial Network (GAN)\cite{GAN}, and Transformer~\citep{NeurIPS17-Transformer}. This lineage of deep learning frameworks has charted remarkable strides in contrast to earlier paradigms of feature engineering. Their prowess spans a spectrum of domains including computer vision, natural language processing, speech, audio processing, and beyond. The amalgamation of these frameworks with vast and diverse domains has culminated in substantial advancements and accomplishments.

Despite the impressive achievements exhibited by these sophisticated representation learning methodologies on extensive datasets across an array of tasks~\citep{CVPR19-StyleGAN,NeurIPS20-3dRcon,ICML14-VisRec}, their effectiveness often falters when confronted with the complexities of real-world situations. A glaring example of this lies in their inability to seamlessly adapt to uncharted territories, such as novel datasets or unfamiliar environments. A pertinent illustration is that of a human detector fine-tuned on indoor images, which proves inadequate when exposed to testing images derived from outdoor contexts. This predicament can be attributed to the concept of domain shift~\citep{PAMI22-DGSurvey,NC18-DASurvey,TKDE22-DGSurvey}, wherein a discernible disparity emerges between the distribution of the training data and that of the testing data. Even slight modifications in the training data can lead to significant performance degradation in numerous deep learning methods~\citep{ICLR19-DNNDegrade,PMLR19-DNNDegrade,CVPR14-DLDegrade,NeurIPS14-DLDegrade} when applied to a target dataset. To enhance the performance of representation learning across domains in real-world scenarios, a multitude of techniques and architectures have been proposed. These approaches aim to tackle the challenges from various angles, collectively striving to address the impact of diverse factors. For instance, in the field of domain adaptation~\citep{NC18-DASurvey}, learning the representations that can reduce the domain gap between source and target is crucial for models to improve the performance~\citep{ICML15-DAN,ICCV15-SDT,NeurIPS18-CDAN}; in the context of domain generalization problem~\citep{TKDE22-DGSurvey,PAMI22-DGSurvey}, it is very important to learn the generalizable representations to improve the robustness of the systems under different application scenarios; with regard to cross-domain few-shot problem, how to enrich the diversity of the support set is one of the pivotal direction to learn the reliable representation to benefit downstream tasks. Inspired by the distinct demands posed by various problems in representation learning across domains, researchers are dedicating increased attention and diligence towards rendering this process more dependable, resilient, equitable, and ethically sound. The objective is to instill trust across diverse domains. In the context of this article, our aim is to deliver a timely and comprehensive survey of recent endeavors that confront the multifaceted challenges associated with establishing "trustworthiness" in representation learning across domains.

 

\subsection{Trustworthy for AI and Representation Learning Across Domains}


In recent years, Artificial Intelligence (AI) has deeply reshaped our daily lives and society. Various AI systems have been deployed in every corner of our like, including transportation, education, public health, security, business, and so forth. We enjoy the economic and social benefits provided AI but also face the challenges come from AI. Because of the complex system design in AI, massive demand for data, the bottleneck of computing power, and the favor of accuracy-only, the AI systems is not as reliable as expected. On the contrary, AI has induced plenty of challenges, including individual and social problems, and reduced the trust of people on it directly or indirectly. For example, many facial recognition systems suffer from the higher error rates for African-Americans, women, and young people~. Another example is the AI systems used in US hospitals to predict which patients would likely need extra medical care heavily favored white patients over black patients. Therefore, many organizations~\citep{ISO-AI,EU-AI} and researchers~\citep{arXiv21-TrustAI,ACMSurvey22-TrustAI} have put much effort to proposed guidelines on how to build a more reliable AI system from different stages of the lifecycle. These endeavor lead the requirement for AI system switch from accuracy to trustworthiness. 

\begin{figure}[t]
\centering
\includegraphics[width=0.5\textwidth]{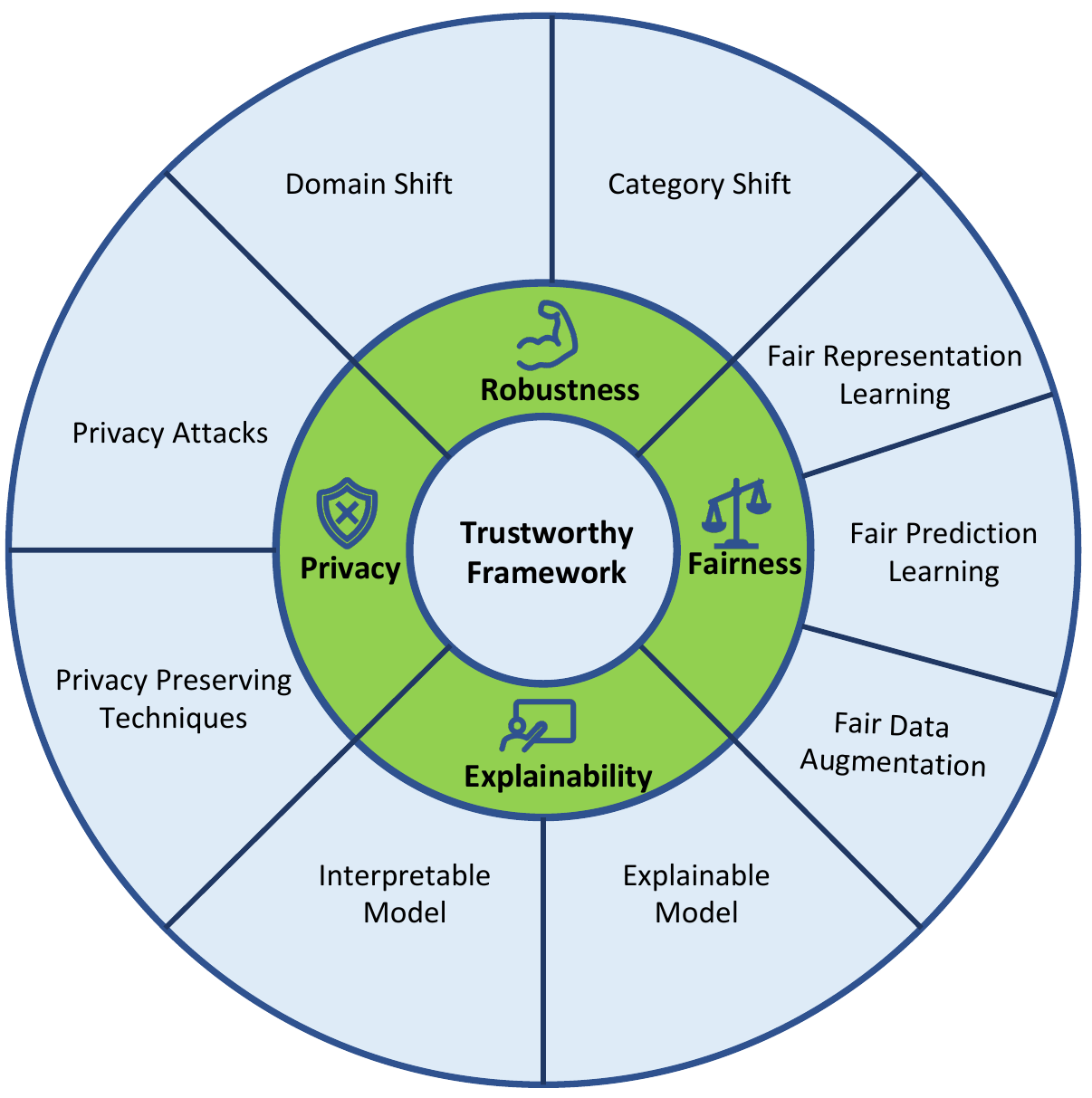}
\caption{Illustration of our proposed trustworthy framework which including four concepts, i.e., robustness, privacy, fairness, explainability.}
\label{overview}
\end{figure}

Drawing inspiration from the achievements of trustworthy AI, the trustworthiness of representation learning across various domains is contingent upon its alignment with the fundamental principles inherent in trustworthy AI. 
As a result, we are convinced of the necessity for a meticulous endeavor to translate the tenets of trustworthy AI into the realm of representation learning across diverse domains. This endeavor seeks to establish a comprehensive framework that harmonizes these principles. Figure~\ref{overview} provides a visual depiction of this framework, highlighting the central components: robustness, explainability, privacy, and fairnes

\subsection{Concepts in Trustworthy Representation Learning Across Domains}
\label{Concepts}
As the trustworthy representation learning across domains is one of the specific cases in trustworthy AI, it inherits the conventional characteristics from trustworthy AI but possesses unique properties. To better illustrate our proposed trustworthy representation learning across domains framework, we will give a brief review of each concept in the framework, including concept definition and research challenges. Please note that these concepts are not orthogonal but closely correlated.


\subsubsection{Robustness}
In the realm of trustworthy AI, robustness entails endowing AI systems with the capability to handle errors—be it data inaccuracies, attacks, or unanticipated inputs—during decision-making, thereby yielding dependable outcomes. This is a foundational concept underpinning the application of trustworthy AI in practical scenarios. The absence of robustness renders an AI system susceptible, where even minor input perturbations can lead to starkly different results. A prime example is the case of an AI system incorrectly classifying a panda image as a gibbon with high confidence due to minute, carefully designed noise~\citep{goodfellow2014explaining}. Within the domain of representation learning across diverse data sources, robustness assumes a distinct connotation. Here, it signifies the capacity of data representations to authentically and accurately depict the underlying distribution spanning different domains. While it's impractical to account for all possible vulnerabilities, our focus chiefly centers on algorithms that facilitate learning data representations enabling consistent performance across domains. Therefore, they could guarantee that AI systems remain robust and keep the trustworthy representation learning intact.



\subsubsection{Explainability}
With the widespread adoption of powerful deep learning models like CNN~\citep{CNN-LeCun} and GAN~\citep{GAN}, contemporary AI systems have become integral to our daily lives, aiding us in decision-making. However, the opaque nature of these complex deep learning models and the inherent black-box property raise concerns. These concerns, voiced by both industry and academia, highlight the lack of internal explanations in many AI systems. The inability to elucidate AI system behaviors has spurred the demand for transparency. Explainability emerges as a pivotal attribute in cultivating trustworthy AI systems. In the context of representation learning across diverse domains, the end-to-end nature of black-box deep learning models often renders learned representations less interpretable. This poses challenges in providing definitive evidence to understand model predictions. Consequently, the focus shifts towards enabling data representations to be more traceable. This traceability empowers researchers to gain deeper insights into learned data representations and offer more comprehensive explanations for model behaviors. For instance, techniques such as saliency maps~\citep{ICLR14-SaliencyMap} can highlight discriminative evidence within images across domains, bolstering the interpretability of models~\citep{CVPRW21-Exp-DG}.


\subsubsection{Privacy}
The objective of privacy is to safeguard the data and model within an AI system from unauthorized exposure. Given that data can encompass sensitive details like familial ties, credit histories, and medical records, the risk of unauthorized access leading to various complications is considerable. In representation learning across domains, the emphasis is on preserving data privacy. In contrast to single-domain scenarios, the exchange of information between multiple domains heightens privacy concerns. The intricate challenge lies in ensuring the security of data when dealing with diverse domains, where the potential for data leakage becomes more pronounced.



\subsubsection{Fairness}
Fairness demands that AI systems exhibit equality, impartiality, and lack of bias towards individuals or groups with varying attributes when making decisions or predictions. Unfortunately, the rapid proliferation of AI systems in our daily routines has exposed instances of human-like discriminatory biases and unjust decisions. For instance, the Association for Computing Machinery (ACM) has advocated for discontinuing the use of facial recognition technologies in both private and government spheres due to glaring biases tied to human characteristics like ethnicity, race, and gender. Additionally, Microsoft's AI-based chatbot on Twitter generated a slew of inappropriate comments, including misogynistic, racist, and anti-Semitic content, underscoring the gravity of the fairness challenge within AI deployment. Therefore, the principle of fairness is an irreplaceable role in the AI system and is vital to eliminate or alleviate the influence of biases that may cause harm to anyone related to the system. In the context of representation learning across domains, 
Fairness is widely applied to different domains so its definition may vary. However, it can basically be defined as the absence of any prejudice or favoritism towards an individual or a group based on their intrinsic or acquired traits in the context of decision-making~\citep{saxena2019how, mehrabi2021survey}. Particularly, fairness refers to the ability to correct data bias or algorithm bias, especially in different domains. In other words, it aims to make the representations and model outputs invariant to sensitive attributes. As for more specific downstream tasks, like cross-domain face recognition, fairness have different goals, such as group fairness~\citep{dwork2012fairness,hardt2016equality}, individual fairness~\citep{dwork2012fairness} and counterfactual fairness~\citep{kusner2017counterfactual}. For example, fairness through unawareness can effectively address group unfairness issues even in new domains, as long as the embeddings used do not contain inherent biased information, such as gender. Additionally, the Lipschitz condition can promote individual fairness in cross-domain areas by ensuring that similar individuals are treated similarly within different domains.

\subsection{Related Surveys}
Representation learning constitutes the foundational challenge in machine learning, and the pervasive integration of artificial intelligence (AI) systems has harnessed the potency of machine learning to achieve notable performance levels, permeating various facets of our societal fabric. In pursuit of fostering trustworthiness, substantial efforts have been channeled, compelling us to address this very concern within representation learning. Particularly, when representation learning encounters cross-domain scenarios, its ability to generalize across varying domains often falters. This realization propelled us to introduce a comprehensive review framework for trustworthy representation learning across domains. While several related surveys exist—spanning the realm of trustworthy AI~\citep{liu2021trustworthy,thiebes2021trustworthy,ACMSurvey22-TrustAI,arXiv21-TrustAI} and representation learning across domains~\citep{PAMI22-DGSurvey,TKDE22-DGSurvey,zhao2020review,kouw2019review,NC18-DASurvey}—none have ventured into the domain we've put forth. Distinguishing itself from these surveys, our endeavor applies universal concepts within trustworthy AI and adapts them to the nuances of representation learning across diverse domains. By bridging this gap, our survey contributes a unique perspective to the intersection of these two crucial spheres.




\subsection{Contributions}
Our contributions are thus summarized as:
\begin{itemize}
\item An Instantiated Trustworthy Framework: We introduce the pioneering trustworthy representation learning across domains framework, outlining its application in various representation learning scenarios across domains. Guided by the principles of trustworthy AI, this framework serves as a roadmap, organizing directions and corresponding technologies in representation learning across domains.
\item Comprehensive Review of Trustworthy Concepts: For each concept embedded within our trustworthy framework, we offer an exhaustive review, delving into the core concepts from fundamental definitions to intricate methodologies.
\item Exploring Future Research Avenues: Through a holistic survey of the diverse representation learning across domains directions encapsulated by our trustworthy framework, we identify promising avenues for future research. By illuminating these potential trajectories, we aim to stimulate further exploration and advancements in the realm of trustworthy representation learning across domains.
\end{itemize}

The remainder of this survey is organised as follows. We first introduce the background of representation learning across domains and the concepts of trustworthy in our framework in Sec.~\ref{background}. Then, Sec.~\ref{robustness},~\ref{privacy},~\ref{fairness},~\ref{explainability} summarizes existing methods for trustworthy representation learning across domains from different concepts in trustworthy. We provide potential directions for our trustworthy framework in Sec.~\ref{future}. Finally, We conclude the survey in Sec.~\ref{conclusion}

\begin{table*}[t]
\centering
\scriptsize
\caption{Illustration of the typical representation learning across domains scenarios. $\mathcal{D}_s$ denotes source domain and $\mathcal{D}_t$ indicates target domain. $\mathcal{D}_t^l$: few labeled target domain. In cross-domain few-shot learning scenario, $\mathcal{D}_t =\{ \mathcal{D}_t^l, \mathcal{D}_t^u$\}, where $\mathcal{D}_t^l$ is few labeled target domain which can be used in training stage and $\mathcal{D}_t^u$ is unseen target domain used to evaluate the learned model.}
\begin{tabular}{l|c|c|c}
\toprule
Scenarios &{Training Data} &{Test Data} &{Condition} \\
\midrule
{Domain Adaptation}~\citep{NC18-DASurvey} &$\mathcal{D}_s$, $\mathcal{D}_t$ &$\mathcal{D}_t$ &$P(X_s, Y_s)\neq P(X_t, Y_s)$ \\
{Domain Generalization~\citep{PAMI22-DGSurvey,TKDE22-DGSurvey}} &$\mathcal{D}_{s_1}, ..., {D}_{s_{n}}$ &$\mathcal{D}_t$ &$P(X_{s_1}, Y_{s_1})\neq ... \neq P(X_{s_n}, Y_{s_n}) \neq  P(X_t, Y_s)$ \\
{Cross-Domain Few-Shot Learning~\citep{ECCV20-CDFSL}} &$\mathcal{D}_s$, $\mathcal{D}_t^l$ &$\mathcal{D}_t^{u}$ &$P(X_s, Y_s) \neq P(X_t^l, Y_t^l)$, $\mathcal{Y}_s \cap \mathcal{Y}_t^l = \emptyset $ \\
\bottomrule
\end{tabular}
\label{RL-Scenarios}
\end{table*}

\section{Background and Concepts}
\label{background}
In this section, we will present the foundational background and essential concepts. The section~\ref{notions} provides the essential notions related to trustworthy representation learning across domains. The sections~\ref{problemDefinition1} and ~\ref{problemDefinition2} present problem definitions with representation learning across domains.

\subsection{Notions and Definitions}\label{notions}
We first give some notions and definitions that will be involved in this article. Let $\mathcal{X}$ be the input data space and $\mathcal{Y}$ be the label space. A domain can be defined as a joint distribution $P(X, Y)$ on $\mathcal{X}\times\mathcal{Y}$. For a specific domain $\mathcal{D}=\{X, Y\}$, we define the $P(X)$ as the marginal distribution on $X$, the $P(Y|X)$ as the conditional probability distribution of $Y$ given $X$, and the $P(X|T)$ as the class conditional distribution of $X$ given $Y$. In general, we can learn the conditional probability distribution $P(Y|X)$ in a supervised way with a specified domain.

In the context of representation learning across domains, we assume to have two kinds of domains, i.e., source domain $\mathcal{D}_s$ which is usually full labeled, and target domain $\mathcal{D}_t$ which is usually unlabeled. Sometimes, the source domain may consist of more than one domain. In general, we hypothesize that there exists a domain sift between source and target domains, i.e., $P(X_s, Y_s)\neq P(X_t, Y_t)$. For source domain $\mathcal{D}_s$, it contains i.i.d. data-label pairs sampled from $P(X_s, Y_s)$, specifically, $\mathcal{D}_s=\{(x^i_s, y^i_s)\}_{i=1}^{n_s}$ with $(x^i_s, y^i_s)\sim P(X_s, Y_s)$. For target domain $\mathcal{D}_t=\{x_t^j\}_{j=1}^{n_t}$, the data are sampled from the marginal distribution $P(X_t)$ due to the label are inaccessible. In general, given the source domain $\mathcal{D}_s$ and target domain $\mathcal{D}_t$, the goal of representation learning across domains is to learn a model $f$ with $\mathcal{D}_s$ or $\mathcal{D}_s$ and $\mathcal{D}_t$ such that the $f$ can generalize well on the target domain $\mathcal{D}_t$.

Problems in representation learning across domains can be organized into several scenarios such as domain adaptation~\citep{TNNLS21-UDA}, domain generalization~\citep{CVPR18-DG-Adv}, and cross-domain few-shot learning~\citep{ECCV20-CDFSL}. See Table~\ref{RL-Scenarios} for an overview. We also consider their combinations with federated learning~\citep{peng2019federated} and continual learning~\citep{MIDL20-CL-DA}. Different application scenarios of representation learning across domains contain different tasks. More details of these application scenarios and related tasks are given below:

\begin{figure}[t]
\centering
\includegraphics[width=\textwidth]{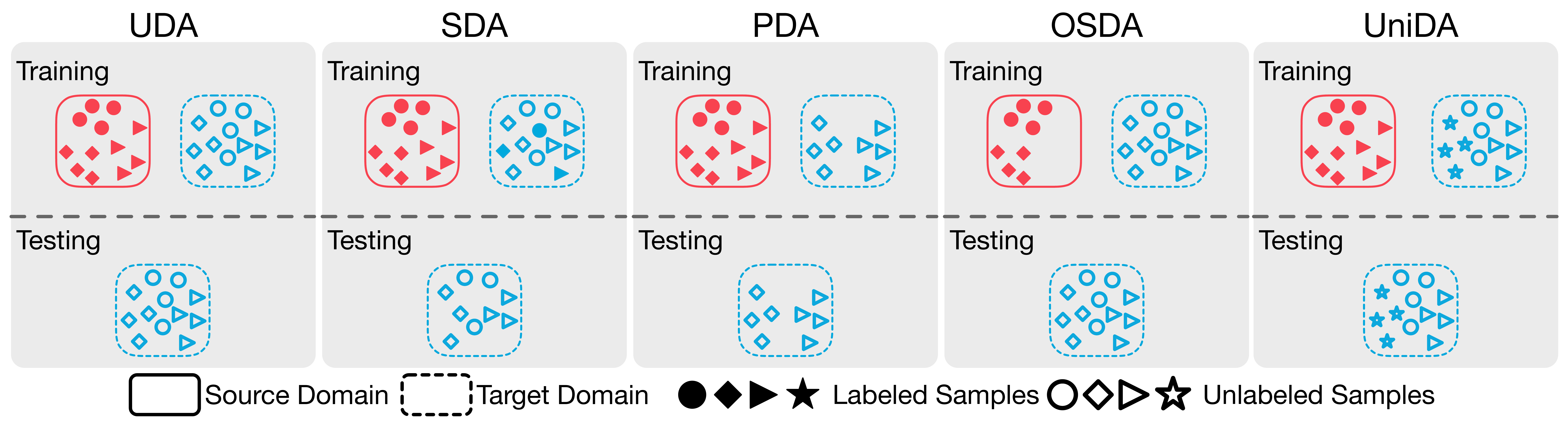}
\caption{Illustration of label space of unsupervised domain adaptation (UDA), Semi-supervised Domain Adaptation (SDA), Partial Domain Adaptation (PDA), Open-Set Domain Adaptation (OSDA), and Universal Domain Adaptation (UniDA). Different colors represent different domains.}
\label{DA_fig}
\end{figure}

\subsection{Domain Adaptation}
\label{problemDefinition1}
\subsubsection{Existing Domain Adaptation Tasks}
Domain Adaptation (DA) is the topic that has been broadly researched over the articles~\citep{CVPR12-DA-UDA,ICML15-DA-UDA,CVPR18-DA-UDA-Adv,CVPR21-DA-SDA}. In general, given two domains, i.e., source domain $\mathcal{D}_s$ which is full labeled and target domain $\mathcal{D}_t$ which is unlabeled or partially labeled. The goal is to develop a model that effectively harnesses data from both the source and target domains to achieve accurate classification of data in the target domain. The key hurdle in domain adaptation arises from the fact that data in these two domains are drawn from dissimilar distributions, compounded by the absence of label information from the target domain. This challenge necessitates innovative strategies to bridge this distribution gap and enable effective classification in the target domain. Learning the data presentations to alleviate the domain shift between two domains and adapting source knowledge to target domain is the major direction to solve the domain adaptation problem. Early, the main goal of the DA is to deal with the domain shift problem where $P(X_s, Y_s) \neq P(X_t, Y_t)$. Recently, DA~\citep{ECCV18-DA-PDA,ICCV17-DA-OSDA,ICCV21-DA-OSDA,CVPR19-DA-UniDA,CVPR21-UniDA} considers a more practical application situation that aims to address both domain shift problem and label shift problem which means that the label spaces of source and target domains are not exactly same ($\mathcal{Y}_s \neq \mathcal{Y}_t$). Many existing domain adaptation (DA) methods fall under the umbrella of Unsupervised Domain Adaptation (UDA)\cite{CVPR12-DA-UDA,ICML15-DA-UDA,CVPR18-DA-UDA-Adv,CVPR22-DA-UDA,CVPR21-DA-UDA,ICCV21-DA-UDA,ECCV16-DA-MomMatch,NeurIPS18-CDAN}. These methods operate under the assumption that the label spaces of both domains are identical, aiming to mitigate the challenges posed by domain shift. On the other hand, Partial Domain Adaptation (PDA)\cite{CVPR21-DA-SDA,ECCV18-DA-PDA,NeurIPS21-DA-PDA,NeurIPS21-DA-PDA1,ECCV20-DA-PDA,CVPR18-DA-PDA} relaxes this assumption by considering scenarios where the source domain's label space encompasses the target domain's label space. Open-Set Domain Adaptation (OSDA)\cite{ICCV17-DA-OSDA,ICCV21-DA-OSDA,ECCV20-OSDA,ICML20-DA-OSDA,CVPR20-DA-OSDA,ICLR19-DA-OSDA} addresses a more realistic application context where the label space of the target domain includes the label space of the source domain. In alignment with real-world needs, Universal Domain Adaptation (UniDA)\cite{CVPR19-DA-UniDA,ECCV2020-DA-UniDA,CVPR21-UniDA,ICCV21-DA-UniDA,CVPR22-DA-UniDA,NeurIPS20-DA-UniDA,ICDM21-DA-ssl} relaxes label space constraints, accommodating private classes unique to each domain and shared common classes between them. These diverse methodologies cater to varying levels of label space alignment, catering to a wide spectrum of DA scenarios. Figure~\ref{DA_fig} illustrates the relationship between source and target label spaces in different DA tasks. We also summarize the difference among different tasks in Table~\ref{DA-tasks}

\begin{table*}[t]
\centering
\scriptsize
\caption{Illustration of the existing domain adaptation (DA) tasks. \textbf{UDA} means Unsupervised Domain Adaptation. \textbf{SDA} means Semi-supervised Domain Adaptation. \textbf{PDA} means Partial Domain Adaptation. \textbf{OSDA} means Open-Set Domain Adaptation. \textbf{UniDA} means Universal Domain Adaptation. $Y_s$ denotes source label space and $Y_t$ indicates target label space. $X_s$ and $X_t$ represent the source and target data.}
\begin{tabular}{l|c|c|c|c|c|c|c}
\hline
\multirow{2}*{Tasks} &\multicolumn{4}{c|}{Label Set} &\multicolumn{3}{c}{Data Set} \\
\cline{2-8}
 &$Y_s = Y_t$ &$Y_t \in Y_s$ &$Y_s \in Y_t$ &$Y_s \cap Y_t \neq \emptyset$ &$X_s$ full labeled &$X_t$ full unlabeled &$X_t$ few labeled \\
\hline
UDA~\citep{TNNLS21-UDA}  &\checkmark &$\times$ &$\times$ &\checkmark &\checkmark &\checkmark &$\times$\\
SDA~\citep{CVPR21-DA-SDA} &\checkmark &$\times$ &$\times$ &\checkmark &\checkmark &$\times$ &\checkmark \\
PDA~\citep{ECCV18-DA-PDA} &$\times$ &\checkmark &$\times$ &\checkmark &\checkmark &\checkmark &$\times$\\
OSDA~\citep{ICCV17-DA-OSDA} &$\times$ &$\times$ &\checkmark &\checkmark &\checkmark &\checkmark &$\times$\\
UniDA~\citep{CVPR19-DA-UniDA} &$\times$ &$\times$ &$\times$ &\checkmark &\checkmark &\checkmark &$\times$\\
\hline
\end{tabular}
\label{DA-tasks}
\end{table*}

\subsubsection{Existing Domain Adaptation Methods}
The ultimate goal lies in domain adaptation is to reduce the domain shift between source and target domains. To address this, Ben-David~\citep{NeurIPS06-DA-risk} proposed the bound for target risk $\epsilon^t(h)$ by using the available source risk $\epsilon^s(h)$ as follows:
\begin{equation}
\label{DA_risk}
\epsilon^t(h) \leq \epsilon^s(h) + d_{\mathcal{H}\Delta\mathcal{H}}(P(X_s),P(X_t)) + \lambda_\mathcal{H},
\end{equation}
where $\epsilon(h)^{s/t}:=\epsilon(h, h^{*s/t})=\mathbf{E}_{x_{s/t}\sim P(X_{s/t})}\left[h(x_{s/t}) \neq h^*(x_{s/t}) \right]$, the $h$ can be any classifier from a hypothesis space $\mathcal{H}$ and the $h^{*s/t}$ is the true classifier on the $P(X_{s/t})$. The $\mathcal{H}\Delta\mathcal{H}$-divergence is defined as:
\begin{equation}
\label{h_divergence}
 d_{\mathcal{H}\Delta\mathcal{H}}(P(X_s),P(X_t)):= \sup_{h,h'\in\mathcal{H}}\left|\epsilon^s(h, h') - \epsilon^t(h, h') \right|,
\end{equation}
and the $\lambda_\mathcal{H}$ denotes the difference in classifier across two domain and is written as:
\begin{equation}
\label{h_divergence}
\lambda_\mathcal{H}:=\inf_{h\in\mathcal{H}}\left|\epsilon^s(h) + \epsilon^t(h) \right|.
\end{equation}
Inspired by this theoretical analysis, the existing DA methods mainly explore the direction of learning domain-invariant feature representations. In general, these methods can be roughly organized into two groups: the moment matching based methods~\citep{ICML15-DAN,arXiv14-DA-MomMatch,ICML19-DA-MomMatch,ICML17-DA-MomMatch}, adversarial learning based methods~\citep{CVPR18-DA-UDA-Adv,ICML15-DA-UDA,NeurIPS18-CDAN,CVPR19-DA-UDA-Adv,CVPR17-DA-UDA-Adv,CVPR20-DA-UDA-Adv}.

\textbf{Moment matching based methods} explicitly measure the discrepancy between the source and target domains on the learned data representations and match the discrepancy with well-defined moment-based distribution measurements~\citep{arXiv21-DA-Survey}. The Maximum Mean Discrepancy (MMD) is one of the most typical measurements defined as:
\begin{equation}
\label{MMD}
MMD(X_s, X_t) = \left\| \frac{1}{n_s}\sum_{i=1}^{n_s}\phi(x_s^i) - \frac{1}{n_t}\sum_{j=1}^{n_t}\phi(x_t^i)  \right\|_{\mathcal{H}},
\end{equation}
where $n_s$ and $n_t$ are the number of data in the source and target domains, $\phi(\cdot)$ is the the kernel-induced feature map that projects from the Reproducing Kernel Hilbert Space (RKHS), and $\left\| \cdot \right\|_{\mathcal{H}}$ is the RKHS norm. Early, Long et al.~\citep{ICML15-DAN} proposed the Deep Adaptation Network (DAN) by designing the multiple kernel variant of maximum mean discrepancies (MK-MMD)~\citep{NeurIPS12-MKMMD} layers to project the hidden representations into an RKHS and match the mean embeddings of source and target domains. The authors of Residual Transfer Networks (RTN)~\citep{NeurIPS16-DA-MomMatch} further extended the DAN with a residual layer and relaxed the restriction on the classifier. Yan et al.~\citep{CVPR17-DA-MomMatch} designed the Weighted Maximum Mean Discrepancy (WMMD) to address the class weight bias across domains problem by providing the class-specific auxiliary weights strategy to alleviate the drawback of MMD on class weight bias. Long et al.~\citep{ICML17-DA-MomMatch} considered to align source and target representations by optimizing the joint distributions of multiple domain-specific layers on a joint maximum mean discrepancy (JMMD) measurement with an adversarial training way. 

Correlation Alignment (CORAL) is another measurement to align the second-order statics (covariance) of source and target domains, which is defined as:
\begin{equation}
\label{CORAL}
CORAL(X_s, X_t)=\frac{1}{4d^2}\left \| C_s - C_t \right \|_F^2,
\end{equation}
where $d$ indicates the feature dimensionality, $\left \|\cdot \right \|_F^2$ represents the squared matrix Frobenius norm, $C_s$ and $C_t$ denotes the feature covariance matrices of $X_s$ and $X_t$, respectively. The authors of Deep Correlation Alignment (Deep CORAL)~\citep{ECCV16-DA-MomMatch} revised the CORAL to nonlinear transformation by incorporating it to the last layer of the deep neural network and designing the corresponding differentiable loss function. Zhuo et al.~\citep{ACMMM17-DA-MomMatch} proposed the Deep Unsupervised Convolutional domain Adaptation (DUCDA), which extended the CORAL not only to fully connected layers in deep neural network but also to convolutional layers, and further improve the generalization of leaned representations. Furthermore, Morerio et al.~\citep{ICLR18-DA-MomMatch} jointly combined the correlation alignment and entropy minimization into a unified framework, i.e., Minimal-Entropy Correlation Alignment (MECA), with geodesic distances to align source and target domains in the learned representations. Chen et al.~\citep{AAAI19-DA-MomMatch} presented the Joint Discriminative Domain Alignment (JDDA) framework, which consists of CORAL loss, instance-based discriminative loss, and center-based based discriminative loss, to learn the discriminative feature representations across domains.

Beside the measurements of MMD and CORAL, there are many other measurements include but not limited to: (1) Zellinger et al.~\citep{ICLR17-DA-MomMatch} proposed a new measurement, i.e., Central Moment Discrepancy (CMD), to match the high order central moments of probability distributions tried to match the higher-order moments by means of order-wise moment differences; (2) Chen et al.~\citep{AAAI20-DA-MomMatch} investigated the higher order statistics, e.g., third and fourth order statistics, for domain alignment by designing the Higher-order Moment Matching (HoMM) framework and further extending the HoMM into RKHS; (3) Shen et al.~\citep{AAAI18-DA-MomMatch} designed the Wasserstein Distance Guided Representation Learning (WDGRL) framework to learning domain-invariant feature representations by minimizing the the estimated Wasserstein distance between source and target domains in an adversarial manner; (4) Jiang et al.~\citep{ACMMM20-DA-MomMatch} presented the Resource Efficient Domain Adaptation (REDA) framework, which distills the transfer knowledge across source and target domains from the top layers to shallower layers in deep neural network by minimizing the Jensen-Shannon Divergence (JSD)~\citep{JSD} between the probability distributions of student and teacher models. (5) Li et al.~\citep{PR18-DA-MoMatch} induced the Adaptive Batch Normalization (AdaBN) model to replace the statistics, e.g., moving average mean and variance, of Batch Normalization (BN) layers trained via source domain to these estimated from the target domain.

\textbf{Adversarial learning based methods} are inspired by the generative adversarial networks (GAN)~\citep{GAN}, which aim to reduce the domain shift across different domains via an adversarial manner. This mechanism has been proved to be the most influential mechanism over others in domain adaptation. Mathematically, the adversarial learning in domain adaptation can be formulated as a minimax optimization problem with two competitive loss objectives: (a) $\epsilon(C)$ on classifier $C$, which is optimized by minimizing the classification error on the source domain; (b) $\epsilon(D, G)$ on domain discriminator $D$ and feature extractor $G$ over source and target domains, which is minimized on $D$ but maximized over $G$:
\begin{equation}
\label{C}
\epsilon(C) = \mathbb{E}_{(x_{s}^{i},y_{s}^{i})\sim\mathcal{D}_{s}}\mathcal{L}_{ce}(C(G(x_s^i),y_{s}^{i}),
\end{equation}
\begin{equation}
\label{DG}
\epsilon(D, G) = -\mathbb{E}_{x_{s}^{i}\sim\mathcal{D}_{s}}\log[D(G(x_s^i))] - \mathbb{E}_{x_{t}^{j}\sim\mathcal{D}_{t}}\log[1-D(G(x_t^j)],
\end{equation}
where $\mathcal{L}_ce$ is the cross-entropy loss. The general formulation of minimax game in adversarial learning is defined as:
\begin{equation}
\begin{array}{ll}
\label{ada}
\min\limits_{C,G}\epsilon(C) - \lambda\epsilon(D,G) \\
\min\limits_{D}\epsilon(D,G),
\end{array}
\end{equation}
where $\lambda$ is the hyper-parameter to balance source risk and domain adversary. One of the earliest adversarial learning based domain adaptation is Domain Adversarial Neural Network (DANN)~\citep{JMLR16-DA-Adv}, which incorporated a minimax game framework into a single feed-forward network with a gradient reversal layer. While the other earliest method, i.e., Adversarial Discriminative Domain Adaptation (ADDA)~\citep{CVPR17-DA-UDA-Adv}, adopted an inverted label GAN loss to divide the optimization process into two independent steps for feature extractor and domain discriminator. Instead of only considering how to align the marginal distribution over soruce and target domains, Long et al.~\citep{NeurIPS18-CDAN} jointly aligned the marginal distribution and conditional distribution with the proposed Conditional Domain Adversarial Networks (CDAN), which includes a conditional discriminator with the concatenate input from feature representation and the corresponding classifier predictions. Zhang et al.~\citep{ICML19-DA-MomMatch} provided a new adversarial domain adaptation margin theory based on the hypothesis-induced discrepancy, and designed a novel measurement, i.e, Margin Disparity Discrepancy (MDD), to facilitate the minimax optimization with rigorous generalization bounds. Xu et al.~\citep{AAAI20-DA-Adv} addressed two problems in adversarial domain adaptation, i.e., insufficient data from two domains to ensure domain-invariance over the whole representation space and hard label for the discriminator to judge real and fake data, and induced the adversarial Domain Adaptation with Domain Mixup (DM-ADA) framework, which includes a mixup version discriminator to guarantee domain-invariance in a more continuous latent space and result in a robustness model across domains. Wei et al.~\citep{NeurIPS21-DA-Adv} alleviated the adverse effort of adversarial domain alignment induced by ignoring how to make the alignment proactively serve the classification task by proposing the Task-oriented Alignment (ToAlign) to decompose the source feature and guide the alignment between source and target features with the prior knowledge from the classification task. Chen et al.~\citep{CVPR22-DA-UDA} tackled the mode collapse problem caused by failing to make full use of predicted discriminative information and design a Discriminator-free Adversarial Learning Network (DALN) which reuses the classifier to replace the discriminator and realizes domain alignment with discriminative feature presentations.

\subsection{Domain Generalization}
\label{problemDefinition2}
\subsubsection{Existing Domain Generalization Tasks}
\begin{figure}[t]
\footnotesize
\centering
\includegraphics[width=\textwidth]{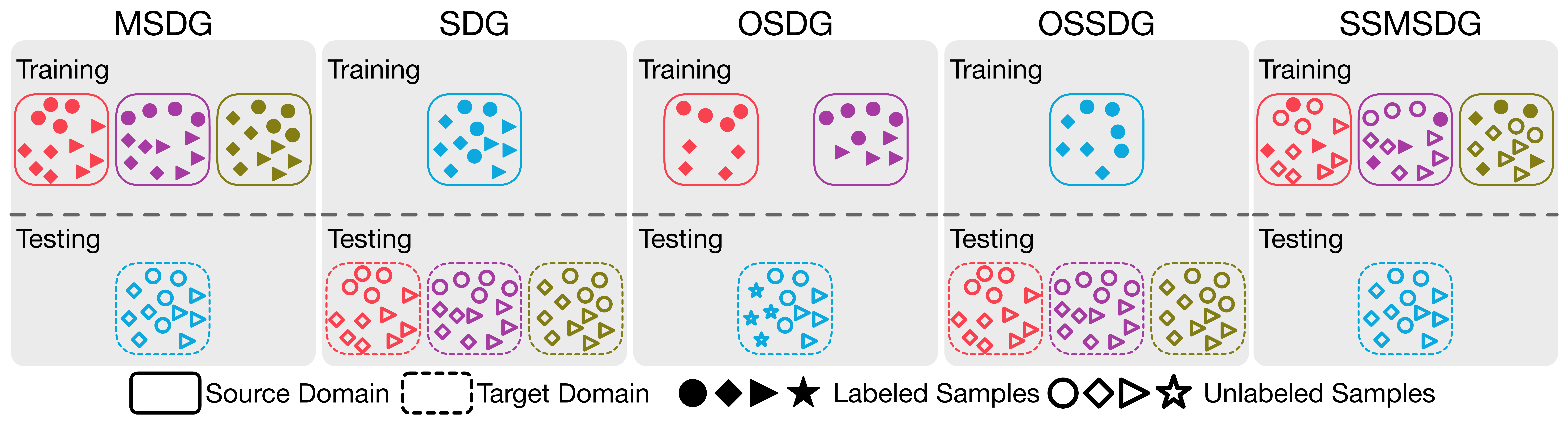}
\caption{Illustration of label space of Multi-Source Domain Generalization (MSDG), Single Domain Generalization (SDG), Open-Set Domain Generalization (OSDG), Open-Set Single Domain Generalization (OSSDG), and Semi-Supervised Multi-Source Domain Generalization (SSMSDG). Different colors represent different domains. As there is no prior knowledge about the target domains, existing MSDG and SDG methods cannot handle unknown classes in target domain.}
\label{DG_fig}
\end{figure}

Domain Generalization (DG) addresses a scenario where the target domain is not accessible during model training. In this setup, multiple fully labeled source domains $\mathcal{D}{s_1}, ..., \mathcal{D}{s_{n_s}}$ are typically available. The key challenge in domain generalization revolves around devising a model that effectively generalizes to an unseen target domain $\mathcal{D}_t$ based on information from diverse source domains. DG is introduced as a solution to both the domain shift problem and the absence of target domain data during model training. Its significance has garnered significant interest from both academia and industry, owing to its potential for real-world applications~\citep{PAMI22-DGSurvey,TKDE22-DGSurvey}. The initial focus of Domain Generalization (DG) was on scenarios involving multiple source domains, each stemming from a distinct distribution. The core objective of DG entails training a model using these diverse source domains, with the aim of enabling it to generalize effectively to an unseen target domain. This specific DG variant is referred to as Multi-Source Domain Generalization (MSDG)~\citep{CVPR18-DG-Adv,NeurIPS21-DG-MSDG,ICLR21-MixStyle,ICLR2021-DG-DomainBed,NeurIPS20-DA-FA-AdvKLD,NeurIPS19-DG-MSDG,ICLR18-DG-MSDG,ICML13-DG-MSDG,ECCV20-DG-MSDG}. In MSDG, the central challenge revolves around training a model that can harness the information from multiple source domains to perform effectively on a target domain that has not been seen during training. Recently, a more challenging and realistic problem, i.e., Single Domain Generalization (SDG)~\citep{CVPR20-DG-SDG,wang2021learning,CVPR21-DG-SDG,CVPR21-SDG-PDEN,CVPR22-DG-SDG,CVPR22-DG-SDG1}, is proposed for DG where only one source domain is accessible for model training. Meanwhile, some researchers further consider more challenging tasks that the label space of source and target domains are not consistent in MSDG and SDG, i.e., Open-Set Domain Generalization (OSDG)~\citep{CVPR21-DG-OSDG} and Open-Set Single Domain Generalization (OSSDG)~\citep{ICLR22-DG-OSDG}. In addition to this, the Semi-Supervised Multi-Source Domain Generalization (SSMSDG)~\citep{NeurIPSW21-DG-SSMSDG} is proposed to address the task where the data in multi-source domain is few labeled. Figure~\ref{DG_fig} illustrates the relationship between source and target label spaces in different DG tasks. We also summarize the difference among different tasks in Table~\ref{DG-tasks}


\begin{table*}[t]
\centering
\footnotesize
\caption{Illustration of the existing domain generalization (DA) tasks. \textbf{MSDG} means  Multi-Source Domain Generalization. \textbf{SDG} means Single Domain Generalization. \textbf{OSDG} Open-Set Domain Generalization \textbf{OSSDG} means Open-Set Single Domain Generalization. \textbf{SSMSDG} means  Semi-Supervised Multi-Source Domain Generalization. $Y_s$ denotes source label space and $Y_t$ indicates target label space. $X_s$ and $X_t$ represent the source and target data.}
\begin{tabular}{l|c|c|c|c|c|c}
\hline
\multirow{2}*{Tasks} &\multicolumn{4}{c|}{Label Set} &\multicolumn{2}{c}{Data Set} \\
\cline{2-7}
 &$Y_s = Y_t$ &$Y_t \in Y_s$ &$Y_s \in Y_t$ &$Y_s \cap Y_t \neq \emptyset$ &$X_s$ full labeled  &$X_s$ few labeled \\
\hline
MSDG~\citep{TNNLS21-UDA}  &\checkmark &$\times$ &$\times$ &\checkmark &\checkmark  &$\times$\\
SDG~\citep{CVPR21-DA-SDA} &\checkmark &$\times$ &$\times$ &\checkmark &\checkmark &$\times$  \\
OSDG~\citep{ECCV18-DA-PDA} &$\times$ &$\times$ &$\times$ &\checkmark &\checkmark  &$\times$ \\
OSSDG~\citep{ICCV17-DA-OSDA} &$\times$ &$\times$ &\checkmark &\checkmark &\checkmark  &$\times$\\
SSMSDG~\citep{CVPR19-DA-UniDA} &\checkmark &$\times$ &$\times$ &\checkmark  &$\times$ &\checkmark\\
\hline
\end{tabular}
\label{DG-tasks}
\end{table*}

\subsubsection{Existing Domain Generalization Methods.}
The primary objective of domain generalization is to train a model using either a single or multiple distinct but interconnected source domains in a way that it demonstrates strong generalization performance on an unseen target domain. The absence of target domain data during model training complicates the estimation of domain shift. In light of this challenge, DG centers on acquiring domain-invariant representations that preserve discriminative qualities on target data. Existing DG methods can be broadly categorized into several groups, including but not limited to: feature alignment-based approaches, meta-learning-based techniques, and data augmentation-based strategies. These methodologies collectively address the task of enabling models to generalize effectively across domains, contributing to the broader landscape of domain generalization research.

\textbf{Feature alignment based methods} aim to align features between different source domains, and hope the learned representations can be invariant to unseen target domain. Along this direction, there are a plethora of methods that adopt different domain alignment strategies such as adversarial learning based and moment matching based, to reduce the domain shift among different source domains. This direction shares the similar motivation to domain adaptation, in some degree, the domain alignment strategies applied in domain adaptation can be used to domain generalization. For example, Li et al.~\citep{CVPR18-DG-Adv} utilized the Maximum Mean Discrepancy (MMD) to minimize the domain shift between source domains, and align the feature distribution to a prior distribution in an adversarial learning manner. Li et al.~\citep{ECCV18-DG-FA-Adv} designed Conditional Invariant Deep Domain Generalization (CIDDG) framework to learn domain-invariant representations via a conditional invariant adversarial network. Zhao et al.~\citep{NeurIPS20-DA-FA-AdvKLD} jointly minimized the KL divergence between the conditional distribution of multiple source domains and applied adversarial learning to match the feature distributions in multiple source domains. Matsuura et al.~\citep{AAAI20-DA-FA-Adv} considered a practical application scenario where domain label information is unavailable, and proposed a framework that can jointly cluster source data into different latent domains and align them under the adversarial learning way. See~\citep{PAMI22-DGSurvey,TKDE22-DGSurvey} for comprehensive surveys on domain generalization)

\textbf{Meta-learning based methods} have drawn increasing attention from the domain generalization community in recent years. The meta-learning is commonly known as learning-to-learn, which attempts to benefit the learning algorithm itself via multiple learning episodes by dividing multiple source data into meta-train $\mathcal{D}_{train}$ and meta-test $\mathcal{D}_{test}$ parts to simulate domain shift. Here we loosely define the model parameters as $\theta$, the overall meta-learning framework can be written as:
\begin{equation}
\begin{array}{ll}
\label{ML1}
\hat{\theta} = \theta^* - \alpha\nabla_{\theta^*}\mathcal{L}_{outer}(\mathcal{D}_{test};\theta^*) \\
\theta^* = \theta - \beta\nabla_{\theta}\mathcal{L}_{inner}(\mathcal{D}_{train};\theta),
\end{array}
\end{equation}
where $\alpha$ and $\beta$ are hyper-parameters, $\mathcal{L}_{inner}(\cdot)$ and $\mathcal{L}_{outer}(\cdot)$ are designed different meta-learning algorithms. The $\mathcal{L}_{outer}(\cdot)$ aims to simulate domain shift and back-propagate the gradients all the way back to the original parameters such that model learned by $\mathcal{L}_{inner}$ can be improved. This learning process is a bi-level optimization and can be formulated as:
\begin{equation}
\label{ML2}
\hat{\theta} = \theta - \alpha\frac{\partial\big(\beta\mathcal{L}_{inner}(\mathcal{D}_{train};\theta) + \mathcal{L}_{outer}(\mathcal{D}_{test};\theta^*)\big)}{\partial \theta}.
\end{equation}
One of the most popular meta-learning frameworks is proposed by Finn~\citep{ICML17-DG-ML}, i.e., Model-Agnostic Meta-Learning (MAML), which separates data into meta-train and meta-test sets, and optimizes the model under meta-train and meta-test tasks to improve the generalization of the learned model. Inspired by this, Li et al.~\citep{AAAI18-DG-ML} are the first to apply MAML to domain generalization problems. They proposed the framework, ie., Meta-Learning for Domain Generalization (MLDG), which divides multiple source domains into uncorrelated meta-train and meta-test 
to mimic the domain shift. The optimization goal is to learn a model with meta-train data that can get low error on meta-test data. In this manner, the learned model can generalize to unseen target domains. Balaji et al.~\citep{NeurIPS18-DG-ML} proposed a regularization based meta learning method (MetaReg) for the classifier, which is different from MLDG meta-learned the entire model. Dou et al.~\citep{NeurIPS19-DG-MSDG} combine two complementary losses which aim to regularize the semantic structure of the feature space, with MAML framework in such a way that the learned representations would keep semantic information across domains and generalize to unseen target domain. Du et al.~\citep{ECCV20-DG-ML} designed Meta Variational Information Bottleneck (MetaVIB) to jointly learn domain-invariant representation and improve the model performance by applying KL Divergence regularization between distributions of latent coding of the samples with the same category but from different domains. Du et al.~\citep{ICLR20-DG-ML} designed an extended version of batch normalization named MetaNorm. They utilize the meat-learning framework to learn adaptive statistics for batch normalization which can improve the reliability and generalizability of learned representations.

\textbf{Data augmentation based methods} usually combine the above-mentioned DG methods such as feature alignment based and meta-learning based, to investigate the benefit brought by data augmentation~\citep{shorten2019survey}. The ultimate goal of data augmentation is to avoid overfitting problem and improve the generalizability of the learned model by generating plenty of diverse data to enrich the original data. Specifically, given the original data $x$, the data augmentation provides the transformation function $Aug(\cdot)$ to generate new data $Aug(x)$ where $x$ and $Aug(x)$ share same semantic information. Early data augmentation techniques focus on basic image manipulations such as: random flip, color space change or transformation, random crop, random rotation, noise injection, kernel filter, random erase, and so on. They can design different $Aug(\cdot)$ by including single or multiple image manipulations. But these $Aug(\cdot)$ are only suitable for certain scenarios. For example, rotation and flip are not applicable for digit classification problems such as "6" and "9" images are easily confused after augmentation. Later, inspired by adversarial attacks~\citep{chakraborty2018adversarial}, adversarial data augmentation~\citep{NeurIPS18-SDG-ADA,NeurIPS20-SDG-MEADA,CVPR21-SDG-PDEN} is proposed, which generates new data by adding sign-flipped gradients back-propagated from classifier to original data. The augmentation process is an iterative way:
\begin{equation}
\label{ADA_max}
x_{t+1} \leftarrow x_t + \eta\nabla_{x_t}\mathcal{L}(\theta;x_t),
\end{equation}
a small number of iterations are needed to produce sufficient perturbations. $\theta$ denotes the parameters in model, and the $\mathcal{L}(\theta; x_t)$ is defined as:
\begin{equation}
\label{ADA_max1}
\mathcal{L}(\theta; x) = \underbrace{\mathcal{L}_{task}(\theta; x)}_{\substack{\text{Task}}} - \underbrace{\alpha\mathcal{L}_{res}(\theta; f)}_{\substack{\text{Restrict}}},
\end{equation}
where $\mathcal{L}_{task}$ is the task loss defined to optimize the model, $\alpha$ is the hyperparameter, $f$ is the feature of original data $x$ extracted by model. The second item in Eq.~\ref{ADA_max1} is noted as $\mathcal{L}_{res}(\theta; f) =\left \|f - f^* \right\|_2^2 $ where $f^*$ is the feature of augmented data $x^*$ which is generated via Eq.~\ref{ADA_max} with original data $x$ as input. This augmentation process can enrich the diversity of the original data while keeping the semantic information. However, these adversarial data augmentation methods fail to simulate the real-word domain shift because the augmented data are visually similar to the original data, e.g., share the same image style. Meanwhile, another direction of data augmentation is proposed~\citep{yue2019domain,AAAI20-DG-DA,ECCV20-DG-MSDG,NeurIPS21-DG-DA}, which considers to enrich the diversity of source domains by synthesizes new domains out-of-source distribution. For example, Zhou et al.~\citep{AAAI20-DG-DA} proposed the Deep Domain-Adversarial Image Generalization (DDAIG) to map the source domains to unseen domains by learning a transformation network in an adversarial manner. Yang et al.~\citep{NeurIPS21-DG-DA} design the Adversarial Teacher-Student Representation Learning (ATSRL) to alternatively learn generalizable representations and generate novel-domain data. Recently, feature-level augmentation has been a fast growing area, motivated by mixup~\citep{ICLR18-Mixup} which linearly interpolates any two data and their labels, respectively, to generate new data, Shu et al.~\citep{CVPR21-DG-OSDG} propose Domain-Augmented Meta-Learning framework to learn generalizable representation by enrich source domains on feature-level via a new designed Dirichlet mixup and label-level though KL divergence. Meanwhile, Zhou et al.~\citep{ICLR21-MixStyle} mix the feature statistics, i.e., mean and standard deviation, of two instances with random ratio to implicitly synthesize new domains.

\subsection{Others}
\textbf{Cross-Domain Few-Shot Learning.} Very recently, many efforts~\citep{ICLR19-FSL,ECCV20-CDFSL} have proven that most of existing few-shot learning~\citep{ACMCS20-FSL-Survey,arXiv22-FSL-Survey} methods fail to exhibit generalization capacity when there is a big domain shift between the source and target domains. To investigate this challenge, Cross-Domain Few-Shot Learning (CDFSL)~\citep{ICLR20-CDFSL,ECCV20-CDFSL} has been proposed. In CDFSL, there are two typical stages involved: 1) meta-learning stage, which utilizes existing source domain $\mathcal{D}_s$ consists of plenty of base category classes to learn a model; 2) meta-testing stage, which adapts the learned model with few-labeled target domain $\mathcal{D}_t^l$ contains a set of novel classes, and hope this adapted model can generalize well to target domain. There is no overlap between the base category classes and the novel classes, i.e., $\mathcal{Y_s}\cap Y_t = \emptyset$, the size of few-labeled target domain is much less than the size of the source domain, and the source and target domains have significant domain shift, e.g., source domain data come from natural images while target domain data come from medical images.

\textbf{Federated Learning.} Federated learning \cite{mcmahan2017communicationefficient,zhang2021survey} aims to maintain a powerful global model via communications with distributed clients without access to their local data. A typical challenge in federated learning is the non-IID data distribution , where the data distributions learnt by different clients are different. 
Formally, A typical Federated Learning paradigm can be formalized as follows: 
Assume we have a set $\textbf{\textit{C}} = \{\mathcal C_{1}, \mathcal C_{2}, ..., \mathcal C_{n}\}$ of $n$ different clients, where each client $\mathcal{C}_{k}$ has its private data $\mathcal{D}_{k}$ with the corresponding task $\mathcal{T}_{k}$, and a global model parameterized by $\boldsymbol{\theta_{g}}$ deployed on a centralized server, which communicates with the clients during their training on their own local data. FL learns a global model that minimizes its risk on each of the client tasks:
\begin{equation}
    \min_{\boldsymbol{\theta_{g}}} \mathbb{E}_{\mathcal{T}_{k} \in \mathcal{T}}\left[\mathcal{L}_{k}(\boldsymbol{\theta_{g}})\right],
    \label{eq1}
\end{equation}

\noindent where $\mathcal{L}_{k}$ is the objective of $\mathcal{T}_{k}$.

\begin{table*}[t]
\footnotesize
\caption{Commonly used benchmark datasets for domain adaption and domain generalization.}
\begin{tabular}{llllll}
\hline 
Dataset    & Domain   & Class & Sample       & Description  &  \\
\hline 
Office-Caltech \cite{qiao2021uncertainty}    & 4        & 10    & 2,533        & Caltech, Amazon, Webcam, DSLR \\
Office-31      \cite{qiao2021uncertainty}    & 3        & 31    & 4,110        & Amazon, Webcam, DSLR \\
PACS           \cite{quinonero2008dataset}    & 4        & 7     & 9,991        & Art, Cartoon, Photos, Sketches \\
VLCS           \cite{kim2021self}    & 4        & 5     & 10,729       & Caltech101, LabelMe, SUN09, VOC2007 \\
Office-Home    \cite{huang2020self}    & 4        & 65    & 15,588       & Art, Clipart, Product, Real \\
Terra Incognita\cite{shi2021gradient}    & 4        & 10    & 24,788       & Wild animal images at locations L100, L38, L43, L46 \\
Rotated MNIST  \cite{wang2021learning}    & 6        & 10    & 70,000       & Digits rotated from $0^{\circ}$ to $90^{\circ}$ with an interval of $15^{\circ}$ \\
DomainNet      \cite{rame2022fishr}    & 6        & 345   & 586,575      & Infograph, Clipart, Painting, Quickdraw, Sketch, Real \\
\hline
\end{tabular}
\label{qdq}
\end{table*}

\subsection{Datasets}
In this section, we summarize several commonly used benchmark datasets in domain adaptation and domain generalization. In table. \ref{qdq}, we first provide an overview of some widely used detests of the image classification task, including Office-Caltech \cite{qiao2021uncertainty}, Office-31      \cite{qiao2021uncertainty}, PACS           \cite{quinonero2008dataset}, VLCS           \cite{kim2021self}, Office-Home    \cite{huang2020self}, Terra Incognita\cite{shi2021gradient}, Rotated MNIST  \cite{wang2021learning}, DomainNet      \cite{rame2022fishr}. Each dataset consists of images from $n$ domains and data from different domains share the same label space.

\section{Robustness in Representation Learning Across Domains}
\label{robustness}
The robustness is the fundamental element in the trustworthy AI framework. In general, it refers to capability of the AI system that remains high performance under different scenarios. It's vital for AI system to establish well reputation of trustworthiness in people and society in real-world application. Ignoring the importance of robustness in AI system will raise massive problems when these systems are deployed to tasks related to safety~\citep{ICII17-AutoDriving}, ethnicity~\citep{news10-face}, healthy~\citep{Science19-healthy}, and so forth. As the results, the earned well reputation will be gradually eroded. In the field of representation learning across domains, the term robustness is to describe the capability of learned representations that lead to a robust model under different circumstances. For example, a self-driving model learned by using urban data should perform well under rural circumstance. A face recognition model trained using asian face data should accurately recognize non-asian face.

In this section, we present a comprehensive review of robustness with domain adaptation, domain generalization, and others, from different aspects such as pseudo-labeling, self-supervision, open-set, and so on. Instead of providing an exhaustive review on every direction, we aim to introduce the mainly focusing lines of the researches and try to lay out how robustness problem in trustworthy representation learning across domains is addressed in recent years. We also demonstrate some potential direction in robustness for future research.

\subsection{Domain Adaptation}
Even though typical domain adaptation (DA) methods have obtained significant achievement in many applications, they still need to improve their robustness in real-world application scenarios. In this section, we mainly focus on three branches to improve the robustness of DA: pseudo-labeling~\citep{ICML17-DA-ps,ICML18-DA-ps,CVPR18-DA-ps,ICCV19-DA-ps,CVPR19-DA-ps,ICML20-DA-ps,NeurIPS21-DA-ps}, self-supervision\cite{arXiv19-DA-ssl,arXiv20-DA-ssl,NeurIPS20-DA-ssl,ICDM21-DA-ssl,ICCV21-DA-ssl}, and open-set~\citep{ECCV18-OSDA,ICCV19-OSDA,CVPR19-OSDA,ECCV20-OSDA,CVPR19-DA-UniDA,ECCV2020-DA-UniDA,NeurIPS20-DA-ssl,ICDM21-DA-ssl,CVPR21-UniDA,ICCV21-DA-UniDA,CVPR22-DA-UniDA}. Most of these methods combine with the typical DA methods, i.e., moment matching based and adversarial learning based, to jointly explore their advantage for robust domain adaptation. Recently, pseudo-labeling has been successfully applied to DA and gains promising performance on many DA tasks. The main challenge for pseudo-labeling base methods is how to fully take advantage of pseudo-labeled target data to benefit the domain alignment instead of misleading domain alignment. Meanwhile, in more real-world situations where the labels in the target domain aren't exactly the same as those in the source domains, researchers have come up with open-set related domain adaptation problems. These are designed to make existing adaptation methods stronger and more adaptable, considering differences in label spaces between different domains. This helps these methods work more robust in practical scenarios. The main goal is to accurately classify target data from the common label space shared with source label space while identify target data from the private label space different from source label space. Very recently, self-supervised learning has shown significant improvement on representation learning by fabricating pretext tasks to excavate supervisory signals from date itself instead of category information. Several endeavors have utilized self-supervised learning to DA problems and achieved significantly performance.

\subsubsection{Robust Domain Adaptation with Pseudo-Labeling}
\begin{figure}[t]
\centering
\includegraphics[width=0.5\textwidth]{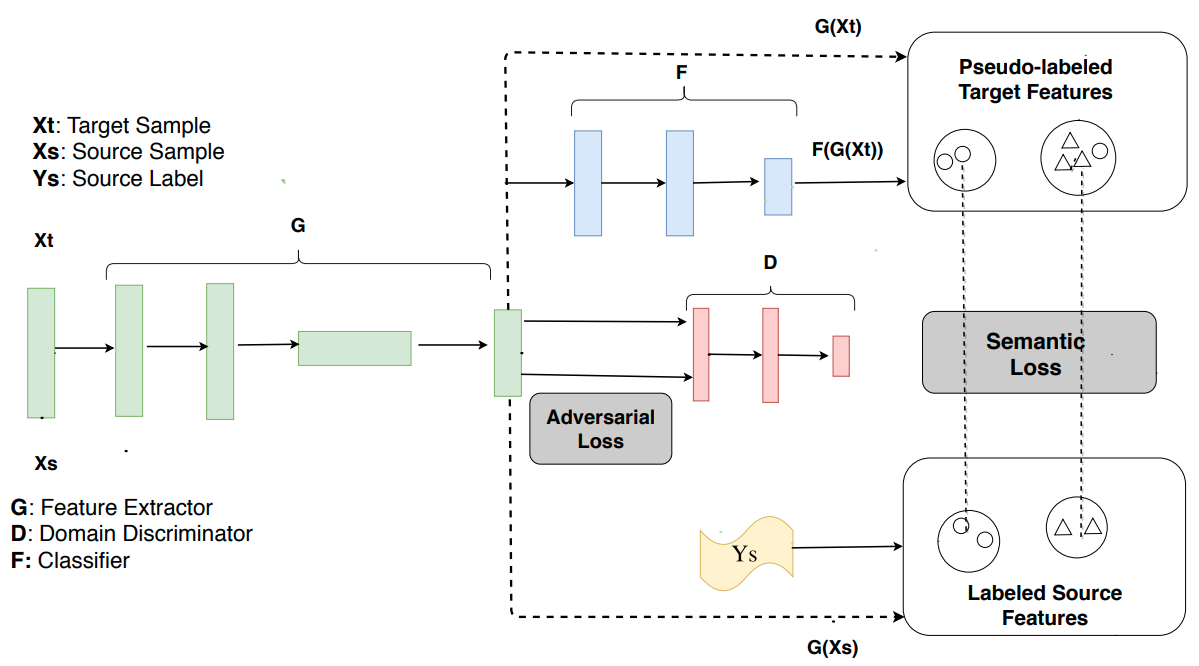}
\caption{The framework of Moving Semantic Transfer Networks (MSTN). (Image from~\citep{ICML18-DA-ps})}
\label{DA_PL}
\end{figure}

\textbf{Pseudo-labeling in DA} operates under the idea that certain target data can be given pseudo-labels and used for training. Saito and colleagues introduced an approach~\citep{ICML17-DA-ps} involving two networks to create pseudo-labels for target data. Then, a third network is trained using these pseudo-labeled target data to develop representations that are specifically geared towards the target domain. Xie et al.~\citep{ICML18-DA-ps} designed the Moving Semantic Transfer Network (MSTN) to realize class-level feature alignment between source and target distribution by assigning pseudo-labels to target data. See Figure~\ref{DA_PL} for an overview. Specifically, the MSTN jointly eliminates the domain shift between source and target domains via adversarial learning manner and align the centroids of source and target via semantic transfer objective. This objective is defined as:
\begin{equation}
\label{MSTN}
\mathcal{L}_{SM}(X_s, Y_s, X_t) = \sum_{k=1}^{K}\phi(C_s^k, C_t^k),
\end{equation}
where $K$ is the number of classes, $\phi(\cdot, \cdot)$,$C_s^k$ and $C_t^k$ denote the centroid of source and target for $k$-th category, respectively. The target centroids are estimated by pseudo-labeled target data. To avoid the error accumulation problem and insufficient categorical information in each batch, they adopted exponential moving average strategy to estimate centroids for source and target domains, which is formulated as:
\begin{equation}
\begin{array}{ll}
\label{MSTN1}
C_s^k \xleftarrow{} \gamma C_s^k + (1 - \gamma)C_{s_{(batch)}}^k\\
C_t^k \xleftarrow{} \gamma C_t^k + (1 - \gamma)C_{t_{(batch)}}^k,
\end{array}
\end{equation}

where $\gamma$ is the hyperparameter controls the contribution of each batch to estimate centroids. $C_{s/t_{(batch)}}^k$ is the mean value of data in source/target batch that belong to $k$-th class. Later, how to safely assign pseudo-label to target data and select pseudo-labeled target data has been explored by many works. For example, Zhang et al.~\citep{CVPR18-DA-ps} designed the Incremental Collaborative and Adversarial Network (iCAN) to iteratively sample a subset of pseudo-labeled target data with a classification or discriminator confidence scores based criterion to retrain the model. Den et al.~\citep{ICCV19-DA-ps} proposed Cluster Alignment with a Teacher (CAT) to investigate the class-wise alignment between source and target distribution by reliably assigning pseudo-labels to target data with a teacher classifier. To further alleviate the error accumulation induced by unreliable pseudo-labels on target data, Chen et al.~\citep{CVPR19-DA-ps} presented the Progressive Feature Alignment Network (PFAN)~\citep{CVPR19-DA-ps}. This approach involves a step-by-step process to choose trustworthy pseudo-labeled target data using cross-domain similarity measurements. It also dynamically adjusts the alignment of source and target class centers to enhance the alignment between different domains. Jiang et al.~\citep{ICML20-DA-ps} addressed the error accumulation problem from sample selection aspect. Specifically, they presented a sampling based implicit method, which utilizes pseudo-labels implicitly to guide data sampling and align the joint distribution between features and labels. Liu et al.~\citep{NeurIPS21-DA-ps} provided a cycle self-training (CST) based method, which improves the reliability of target pseudo-labels by inducing the Tsalli entropy regularization, and promotes the generalizability of pseudo-labels across domains.

\begin{figure}[t]
\centering
\includegraphics[width=0.45\textwidth]{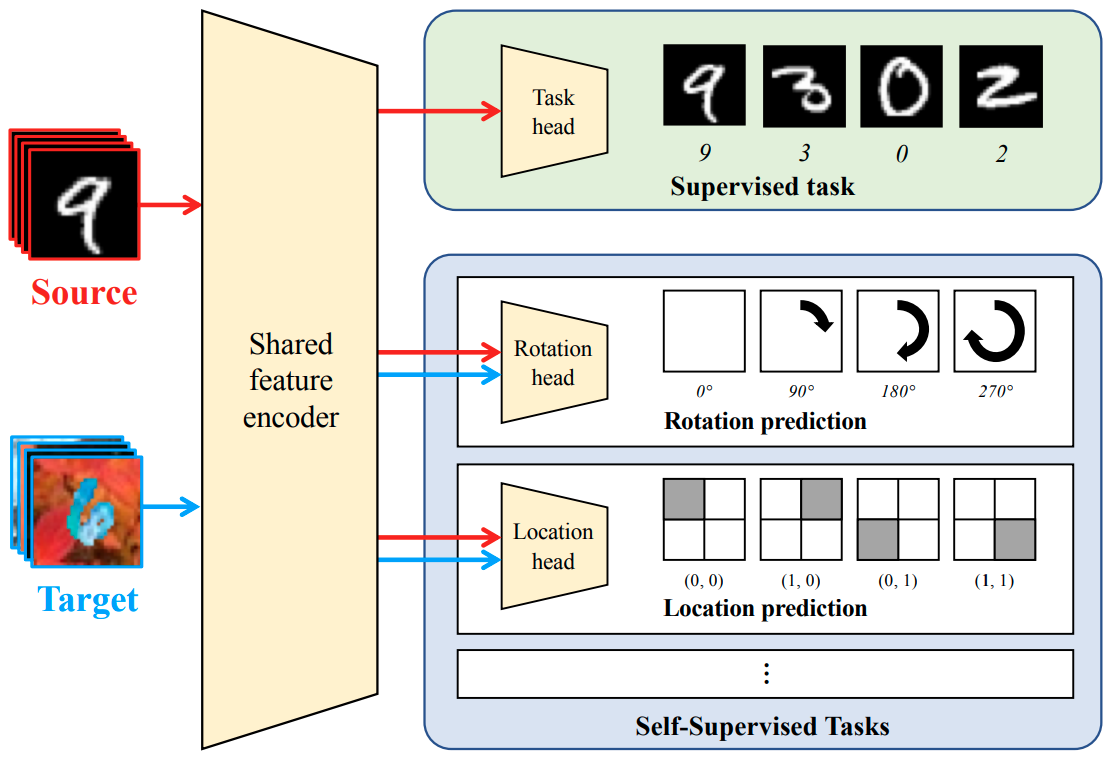}
\caption{The self-supervised learning based framework for unsupervised domain adaptation. (Image from~\citep{arXiv19-DA-ssl})}
\label{DA_SSL1}
\end{figure}

\begin{figure}[t]
\centering
\includegraphics[width=0.9\textwidth]{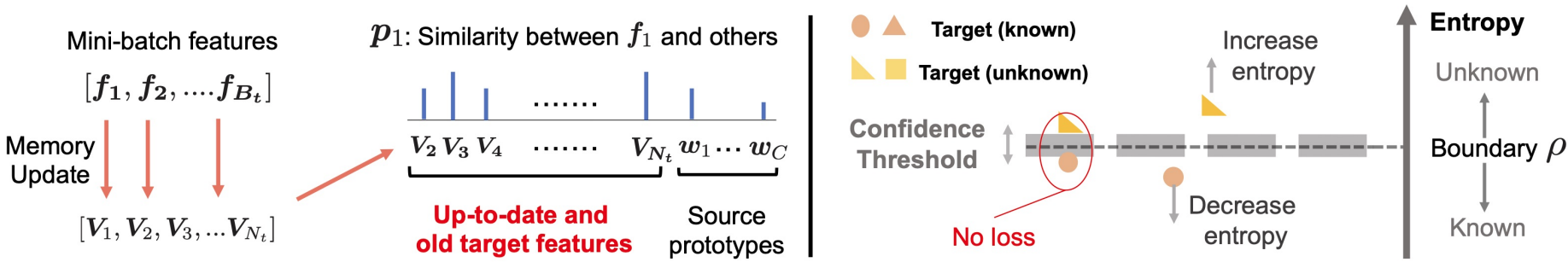}
\caption{The illustration of Domain Adaptative Neighborhood Clustering via Entropy Optimization (DANCE). \textbf{Left:} Similarity distribution calculation in neighborhood clustering. \textbf{Right:} An overview of the entropy separation loss. (Image from~\citep{NeurIPS20-DA-ssl})}
\label{DA_SSL2}
\end{figure}

\subsubsection{Robust Domain Adaptation with Self-Supervision}
\textbf{self-supervision in DA} shares the spirit of the self-supervised learning~\citep{PAMI20-SelfSL-Survey}, and incorporates the self-supervised learning pretext tasks, e.g., image rotation~\citep{rotation}, patch location prediction~\citep{noroozi2016unsupervised}, and instance discrimination~\citep{ID} (used for contrastive learning), into existing DA methods to improve their robustness. Early, Sun et al.~\citep{arXiv19-DA-ssl} added auxiliary self-supervised tasks such as rotation prediction, flip prediction, and patch location prediction, for source and target domains. See Figure~\ref{DA_SSL1} for details. In this framework, the over objective is written as:
\begin{equation}
\label{MSTN}
\min_{\theta, h_k,k=0...k } \mathcal{L}_0(X_s;\theta, h_0) + \sum_{k=1}^K\mathcal{L}_k(X_s, X_t;\theta, h_k),
\end{equation}
where $\theta$ is parameters of feature encoder, $h_k$ denotes the parameters of the $k$-th head task model, e.g., rotation prediction and location prediction task models, $h_0$ is the actual prediction task, and $\mathcal{L}_k$ is the predefined loss function for $k$-th head task model and feature encoder. By optimizing the objective, the model will encourage both domains to align together along the direction related to tasks. Recently, inspired by the success of contrastive learning in self-supervised learning, Park et al.~\citep{arXiv20-DA-ssl} proposed the Joint Contrastive Learning (JCL) to learn discriminative feature representations, which investigates the mutual information between features and labels under the contrastive learning framework. Different from previous DA methods, Saito et al.~\citep{NeurIPS20-DA-ssl} proposed a universally applicable domain adaptation framework, named as Domain Adaptative Neighborhood Cluster via Entropy Optimization (DANCE), to tackle arbitrary category shift. See Figure~\ref{DA_SSL2} for an overview. The over optimization function is defined as:
\begin{equation}
\label{DANCE}
\mathcal{L}_{all} = \mathcal{L}_{ce} + \lambda(\mathcal{L}_{nc} + \mathcal{L}_{es}),
\end{equation}
where $\lambda$ is the hyperparamter, $\mathcal{L}_{ce}$ is the cross-entropy loss function, $\mathcal{L}_{nc}$ denotes the proposed neighborhood clustering loss function (See Figure~\ref{DA_SSL2} left part for an overview), and $\mathcal{L}_{es}$ represents the well-designed entropy separation loss function (See Figure~\ref{DA_SSL2} right part for an overview). The goal of $\mathcal{L}_{nc}$ is to learn discriminative and generalizable representation for target domain, which defined as:
\begin{equation}
\label{DANCE1}
\mathcal{L}_{nc} = -\frac{1}{\left| B_t\right|}\sum_{i\in B_t}\sum_{j=1, j\neq i}^{N_t + K}p_{i,j}log(p_{i,j}),
\end{equation}
where $B_t$ is the sets of target samples' indices in the batch, $N_t$ is the number of target data, $K$ is the number of source classes, and $p_{i,j}$ denotes the probability of the target feature $f_i$ is the neighbor of $j$-th item in memory bank $V$, which stores all target features and prototype weight vectors. The $p_{i,j}$ is computed by:
\begin{equation}
\label{DANCE2}
p_{i,j}=\frac{\exp(F_j f_i /\tau)}{\sum_{j=1,j\neq i}^{N_t + K}\exp(F_j f_i / \tau)}.
\end{equation}
In order to separate target date into known classes and unknown classes, the entropy separation loss function $\mathcal{L}_{es}$ is formulated as:
\begin{equation}
\label{DANCE3}
\mathcal{L}_{es}=\frac{1}{\left | B_t \right |} \sum_{i\in B_t} \mathcal{L}_{es}(p_i), \mathcal{L}_{es}(p_i) = 
\begin{cases}
-\left | H(p_i) - \rho\right |, & \left| H(p_i) - \rho \right | > m , \\
0, & otherwise.   
\end{cases}
\end{equation}
In $\mathcal{L}_{es}$, $p_i$ is the output of classifier for target feature $f_i$, $H(p_i)$ is the entropy value, $\phi$ is threshold to separate target data into known and unknown classes defined as $\frac{\log{(K)}}{2}$. The over all functions aim to learn the representations for target domain by incorporating contrastive learning to dig the similarity of target data and source class prototypes. Zhu et al.~\citep{ICDM21-DA-ssl} further extended DANCE by providing different neighbors for different target data to excavate the similarity under the constrastive learning framework and improve the consistency of classifier for each target under different augmentations. Kim et al.~\citep{ICCV21-DA-ssl} proposed Cross-Domain Self-supervision (CDS) to learn discriminative and domain-invariant representations by jointly mining the similarity of source and target data in-domain and cross-domain under the contrastive learning.

\subsubsection{Robust Domain Adaptation with Open-Set}
\textbf{Open-set in DA} aims to address the practical problem where source and target have different label spaces, and promote the robustness of DA in real-world application. In general, open-set in DA considers two kinds of tasks: the first one is Open-Set Domain Adaptation (OSDA), which assumes that the target label space contains unknown classes for source label space. The later one is Universal Domain Adaptation (UniDA), which removes the restrictions on both label spaces, which allows source and target domains not only have their private label spaces but also share a common label space.

\begin{figure}[t]
\centering
\includegraphics[width=0.9\textwidth]{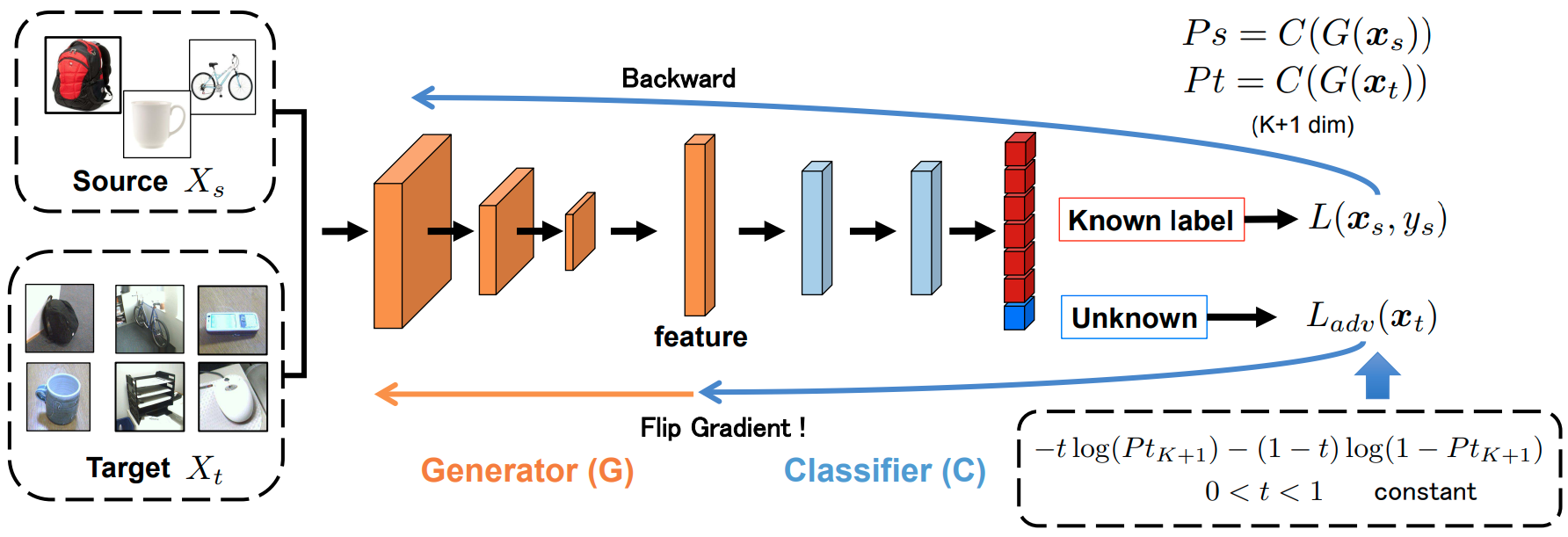}
\caption{The adversarial learning based framework for open-set doamin adaptation.(Image from~\citep{ECCV18-OSDA})}
\label{DA_OS1_}
\end{figure}

For OSDA, the early method~\citep{ECCV18-OSDA} is based on adversarial learning framework, which aligns target data with source domain or reject them as the unknown target classes which don't belong to source label space. See Figure~\ref{DA_OS1_} for an overview. They designed the classifier $C$ to classify data into $K+1$ classes where $K$ is the number of classes in source label space while $1$ indicates the target data from unknown classes. The overall objective function is defined as:
\begin{equation}
\begin{array}{ll}
\label{DA_OS}
\min\limits_{C} \mathcal{L}_{ce}(x_s, y_s) + \mathcal{L}_{adv}(x_t),\\
\min\limits_{G} \mathcal{L}_{ce}(x_s, y_s) - \mathcal{L}_{adv}(x_t),
\end{array}
\end{equation}
where $\mathcal{L}_{ce}(\cdot, \cdot)$ is the cross-entropy loss, and $\mathcal{L}_{adv}(\cdot)$ is a binary cross-entropy loss formulated as:
\begin{equation}
\label{DA_OS1}
\mathcal{L}_{adv}(x_t) = -\alpha\log{(p(y= K+1|x_t)} - (1-\alpha)\log{(1-p(y=K+1|x_t))},
\end{equation}
where $p(y=K+1|x_t)$ denotes the probability of $x_t$ being classified as the unknown class. $\alpha$ is designed to train the classifier to output $p(y=K+1|x_t)=\alpha$, which is set to $0.5$. Under the adversarial optimization process, the learned representations would gradually separate the target data into known and know classes. For these known classes data, they will be aligned to source data. Later, Feng et al.~\citep{ICCV19-OSDA} and Liu et al.~\citep{CVPR19-OSDA} shared the similar idea of separating target data into known and unknown classes, and aligning the target data from known classes to source data. Both of them are based on adversarial learning framework. In one approach, researchers delved into the semantic structure of both the source and target domains. They achieved this by aligning the centroid of target data from recognized classes with their corresponding centroid from the source domain. This strategy also involved pushing target data from unfamiliar classes away from the decision boundary. In another approach, a coarse-to-fine weighting mechanism was introduced. This technique gradually separated target data into known and unknown classes. Then, they devised a weighted adaptation process to align the distributions of target and source data, refining the alignment step by step. Recently, instead of using adversarial learning, Bucci et al.~\citep{ECCV20-OSDA} proposed a self-supervision based framework, which uses rotation recognition as the pretext task, to separate known and unknown target data and align the distributions of source and known target. To explore the useful information in target data belong to unseen classes, Jiang et al.~\citep{ICCV21-DA-OSDA} proposed a novel problem, termed as Semantic Recovery Open-Set Domain Adaptation (SR-OSDA), which based on the OSDA problem definition and further provides the semantic attribute annotation on source domain. They designed a framework to simultaneously learn domain-invariant representation and build visual-semantic projection to restore the lacking of semantic attribute of target unknown classes.

\begin{figure}[t]
\centering
\includegraphics[width=0.9\textwidth]{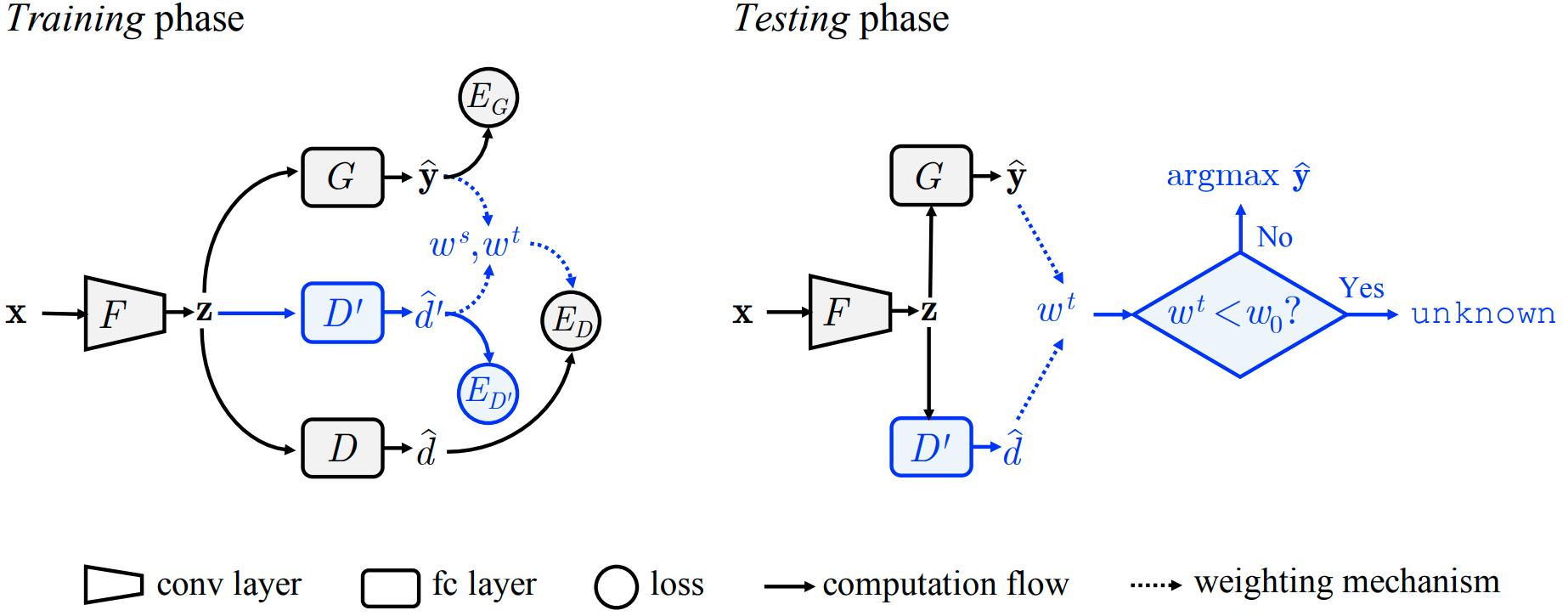}
\caption{The framework of Universal Adaptation Network (UAN). $F$ is the feature extractor, $G$ indicates the classifier, $D$ represents the adversarial domain discriminator, and $D'$ denotes the non-adversarial domain discriminator which used to compute the domain similarity value. the (Image from~\citep{CVPR19-DA-UniDA})}
\label{DA_UniDA}
\end{figure}

For UniDA, a more practical and challenging problem than OSDA, the first UniDA method is Universal Adaptation Network (UAN)~\citep{CVPR19-DA-UniDA}, which employs a metric consists of entropy and domain similarity values to measure the transferability of each target data and distinguish target data from common classes and those from private classes. See Figure~\ref{DA_UniDA} for an overview. In the training phase, the overall optimization objective is defined in an adversarial manner:
\begin{equation}
\begin{array}{cc}
\label{UniDA}
\max\limits_{D}\min\limits_{F,G}E_G - \lambda E_D, \\
\min\limits_{D'}E_{D'},
\end{array}
\end{equation}
where $E_G$, $E_D$, and $E_{D'}$ are the error for classifier $G$, adversarial domain discriminator $D$, and non-adversarial domain discriminator $D'$. They are defined as:
\begin{equation}
\begin{array}{lll}
\label{UniDA1}
E_G = \mathbb{E}_{(x_s,y_s)\sim \mathcal{D}_s}\mathcal{L}_{ce}\big(G\big(F(x_s)\big), y_s\big), \\
E_{D}=-\mathbb{E}_{x_s\sim\mathcal{D}_s}\omega_s(x_s)\log{D\big( F(x_s)\big)} -\mathbb{E}_{x_t\sim\mathcal{D}_t}\omega_t(x_t)\log{\big(1 - D\big( F(x_t)\big)\big)}, \\
E_{D'}=-\mathbb{E}_{x_s\sim\mathcal{D}_s}\log{D'\big( F(x_s)\big)} -\mathbb{E}_{x_t\sim\mathcal{D}_t}\log{\big(1 - D'\big( F(x_t)\big)\big)},
\end{array}
\end{equation}
where $\omega_s(x_s)$ and $\omega_t(x_t)$ are two sample-level transferability criterions for source and target data, that used to constrain the $D$ to distinguish the source and target data in the common label space. They are defined as:
\begin{equation}
\begin{array}{cc}
\label{UniDA2}
\omega_s(x_s) = \frac{H\big(G\big(F(x_s) \big) \big)}{\log{(C_s)}} - D'\big(F(x_s) \big)  \\
\omega_t(x_t) = D'\big(F(x_t)\big) - \frac{H\big(G\big(F(x_t)\big)\big)}{\log{(C_s)}},
\end{array}
\end{equation}
where $H(\cdot)$ is the entropy value. By utilizing these two criterions, UAN can disentangle source and target data into common classes and their private classes. In this way, the category gap and domain shift are jointly reduced. However, the proposed metric is less robust when entropy value is less reliable and domain similarity value is overconfident. Fu et al.~\citep{ECCV2020-DA-UniDA} proposed Calibrate Multiple Uncertainties (CMU) to offer a more robust metric, which consists of entropy, confidence, and consistency from an ensemble framework. Different from UAN and CMU, Saito et al.~\citep{NeurIPS20-DA-ssl} deal with the UniDA problem by utilizing constrastive learning to learn generalizable and discriminative representations and using entropy value of classifier to separate target data into known and unknown classes. Inspired by~\citep{NeurIPS20-DA-ssl}, Zhu et al.~\citep{ICDM21-DA-ssl} proposed a more robust framework to learn better representations and improve the robustness of entropy value of classier by inducing consistency regularization on target data under different data augmentation. Recently, Li et al.~\citep{CVPR21-UniDA} investigate discriminative clusters in source and target domains to discover the intrinsic structure of target domain and separate target data into known classes and unknown classes by exploiting domain consensus knowledge. Saito et al~\citep{ICCV21-DA-UniDA} proposed One-vs-All Network (OVANet) to adaptively learn the threshold to separate target data into known and unknown classes by exploring the inter-class distance between source categories. However above methods ignore the importance of utilizing the intrinsic manifold structure relationship between source and target domains, which could lead their model less robust to real-world application. To address this issue, Chen et al.~\citep{CVPR22-DA-UniDA} proposed a global joint local learning framework, which includes geometric adversarial learning for domain alignment, subgraph-level contrastive learning for local region aggregation, and universal incremental classifier for automatic unknown threshold learning, to better align target data from known classes to source data while separate these from target unknown classes data. 


\subsection{Domain Generalization}
Compared with domain adaptation, domain Generalization (DG) is more difficult to learn a robust model that can generalize to target domain. Because it has to not only address the domain shift between source and target domains but also face the absence of target domain in training stage. Although plenty of existing DG methods have achieved significant performance, there is still a long way to realize robustness for domain generalization in real-world application. In this section, we mainly focus on two branches to review how to learn robust representations in DG under the real application scenarios: single source domain based methods and open-set related methods. Different from the assumption in typical domain generation methods that these are multiple source domain can be used to learn the model, single source domain base methods consider a more challenging and realistic problem, which assumes only one labeled source domains is available for training and many unseen target domains for test.  Very recently, open-set related tasks have been proposed under the domain generalization, which require the learned model can accurately classify each target data into either a known class in source label space or an unknown class.

\begin{figure}[t]
\centering
\includegraphics[width=0.5\textwidth]{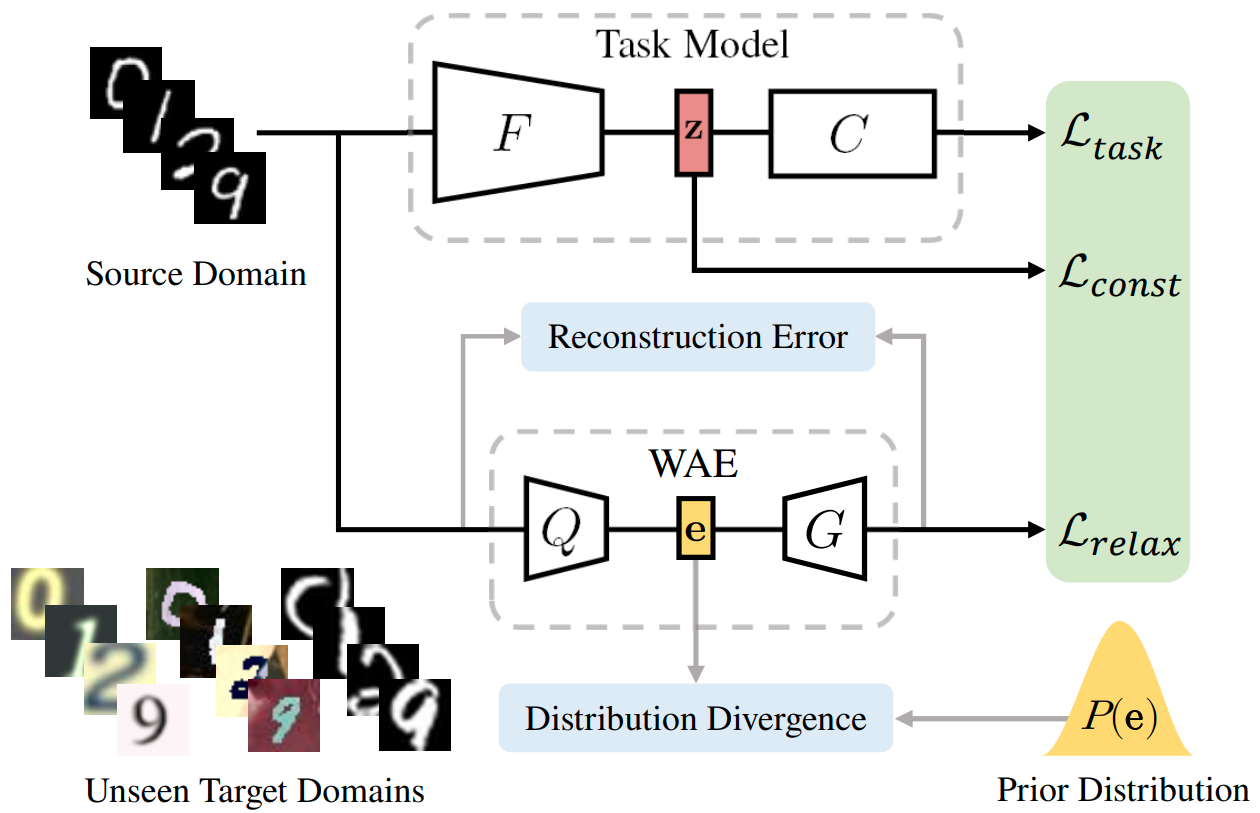}
\caption{The framework of Meta-Learning based Adversarial Domain Augmentation (M-ADA). $F$ is the feature extractor, $C$ represents the classifier, $Q$ and $G$ denote encoder and decoder, respectively. (Image from~\citep{CVPR20-DG-SDG})}
\label{DG_DA}
\end{figure}

\subsubsection{Robust Domain Generalization with Single Source Domain} \textbf{Single source domain in DG} addresses a practical application scenario in DG that only one source domain is available to train the model. This problem is also well-known as Single Domain Generalization (SDG), which is the worst-case domain generalization scenario. One of the earliest SDG methods, named as Meta-Learning based Adversarial Domain Augmentation (M-ADA), is proposed by Qiao et al.~\citep{CVPR20-DG-SDG}, which adopts the adversarial data augmentation (ADA)~\citep{NeurIPS18-SDG-ADA} into meta-learning framework and utilizes an auxiliary Wasserstein autoencoder to help ADA generate more challenging data to enrich the diversity of source domain. See Figure~\ref{DG_DA} for details. They relaxed the ADA objective in Eq.~\ref{DA_SSL1} to:
\begin{equation}
\label{SDG}
\mathcal{L}(\theta; x) = \underbrace{\mathcal{L}_{task}(\theta; x)}_{\substack{\text{Task}}} - \underbrace{\alpha\mathcal{L}_{res}(\theta; f)}_{\substack{\text{Restrict}}} + \underbrace{\beta\mathcal{L}_{rel}(\psi; x)}_{\substack{\text{Relax}}},
\end{equation}
where $\psi$ denotes the parameters in encoder $Q$ and decoder$G$. $\mathcal{L}_{rel}$ ensures large domain transportation defined as:
\begin{equation}
\label{SDG}
\mathcal{L}_{rel}(\psi;x) = \left\|x - G(Q(x)) \right\|,
\end{equation}
here $Q$ and $G$ are pre-trained to capture the distribution of the source domain. Adopting the same augmentation process as ADA in Eq.~\ref{ADA_max}, the M-ADA will generate diverse data to enrich source distribution. They further jointed source data and its augmented version into a meta learning framework to learn the model that can generalize to unseen target domain. Zhao et al.~\citep{NeurIPS20-SDG-MEADA} proposed Maximum-Entropy Adversarial Data Augmentation (MEADA) to generate effective new data distributions consisting of "hard" data that are significantly different from source data by adding the maximum-entropy regularizer to ADA. Fan et al.~\citep{CVPR21-DG-SDG} explored the impact of the statistics, i.e., standardization and rescaling statistics, of normalization layers to improve the generalization of the model. They proposed the Adaptive Standardization and Rescaling Normalization (ASR-Norm) to complete the ADA for data augmentation by learning related statistics in network. Wang et al.~\citep{wang2021learning} proposed a style-complement module to improve the capability of model on generating new data that are out of source distribution. In their framework, a mutumal information based minmax game is played to enhance the ability of style-complement module on data generation and promote the generalizability of model on unseen target domain. Li et al.~\citep{CVPR21-SDG-PDEN} designed Progressive Domain Expansion Network (PDNE) to jointly generate new domains in a progressive manner via the domain expansion subnetwork and learn domain-invariant representation though a contrastive learning based subnetwork. Cugu et al.~\citep{CVPR22-DG-SDG1} apply a diverse set of visual corruptions to enrich source distribution and propose an attention consistency loss to align the class activation maps between original data and the corrupted version. Different from above data augmentation based methods, Wan et al.~\citep{CVPR22-DG-SDG} address the SDG problem from the aspect of model architecture, they proposed a new convolutional model to extract unbiased universal feature for unseen target data. 

\begin{figure}[t]
\centering
\includegraphics[width=0.85\textwidth]{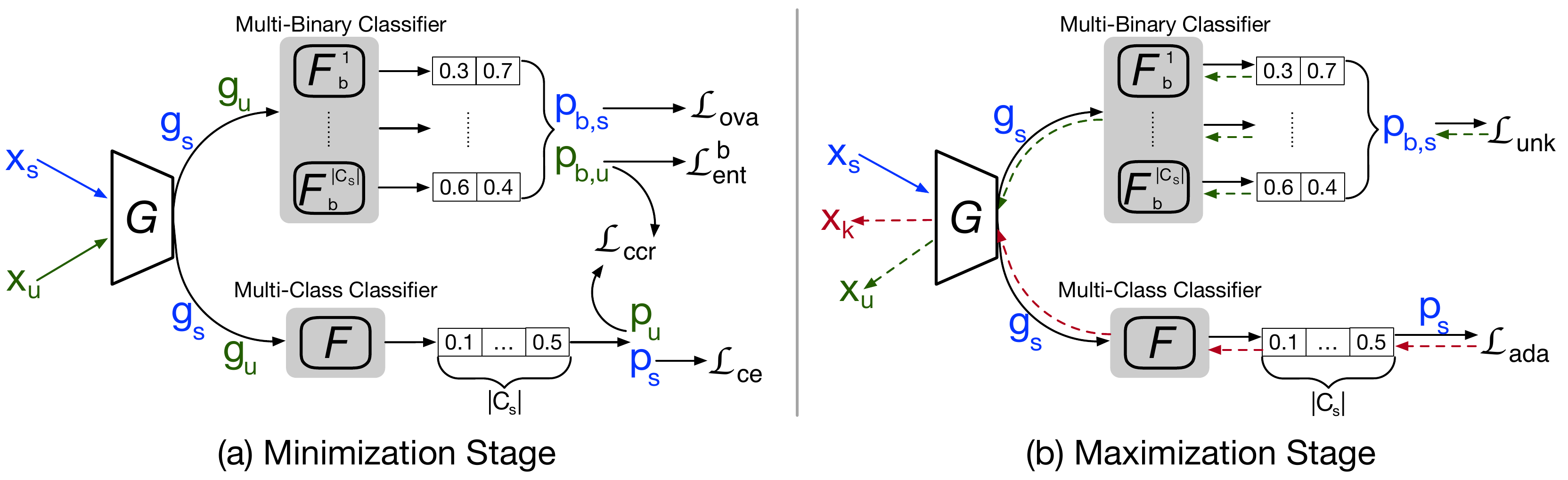}
\caption{The framework of Cross-Match. $G$ is the feature extractor, $F$ represents the classifier, multi-binary classifier is denoted as $F_b=\{F_b^1, \ldots, F_b^{|\mathcal{C}_s|}\}$ where $|C_s|$ is the number of classes in source label space. (Image from~\citep{ICLR22-DG-OSDG})}
\label{OSSDG}
\end{figure}

\subsubsection{Robust Domain Generalization with Open-Set} Open-set in DG simulates the real-world application scenarios that unseen target domain has domain shift and a different label space from the source domains. This practical problem is difficult for existing typical DG methods. Because their learned domain-invariant representations can generalize to target domain is based on the assumption that source and target domains share the same label space. When target domain contains data not belong to source label space, the learned representation would likely fail to generalize to target domain. Very recently, there are two works~\citep{CVPR21-DG-OSDG,ICLR22-DG-OSDG} try to tackle the open-set problem from different aspects. Shu et al.~\citep{CVPR21-DG-OSDG} address the open-set problem under the assumption that there are multiple source domains available for model training. They proposed Domain-Augmented Meta-Learning (DAML) to learn generalizable representation by enriching source domains on feature-level via a new designed Dirichlet mixup and label-level though KL divergence. While, Zhu et al.~\citep{ICLR22-DG-OSDG} tackle the open-set problem under the assumption that there only one source domain accessible for model training. They proposed CrossMatch framework, which jointly generates auxiliary data to simulate target unknown classes data and applies consistency regularization on generated auxiliary data between multi-binary classifiers and the model to improve the capability of learned model on unknown class identification. See Figure~\ref{OSSDG} for an overview. One of their major contribution is to generate auxiliary data that potentially out of source distribution to simulate unknown classes data. They followed the same strategy in adversarial data augmentation~\citep{NeurIPS18-SDG-ADA} to generate auxiliary data. Similar as the Eq.~\ref{ADA_max1}, their augmentation process is written as:
\begin{equation}
\label{G_U_eq}
x^{t+1}_u \leftarrow x^t_u + \eta\nabla_{x^t_u}\{\mathcal{L}_{unk}(x_u^t) + \gamma\mathcal{L}_{res}(\theta_g; x_u^t, x_s)\},
\end{equation}
where $\mathcal{L}_{res}$ is the worst-case guarantee defined in Eq.~\ref{ADA_max1}. $\mathcal{L}_{unk}$ is their designed auxiliary data generalization objective which is formulated as:
\begin{equation}
\label{Unk_loss}
\mathcal{L}_{unk}(x_s,y_s) = -\log(p_{b,s}^{y_s}(t=0|x_s)) + \sum\limits_{i\neq y_s}^k\log{p_{b,s}^i(t=1|x_s)},
\end{equation}
where $p_{b,s}^{y_s}(t=0|x_s)$ and $p_{b,s}^{y_s}(t=1|x_s)$ represent the probabilities of being $y_s$-th class and other classes from the $y_s$-th binary classifier, respectively. By maximizing $\mathcal{L}_{unk}$ at the maximization stage, it encourages the model to generate new samples that all binary classifiers in $F_b$ predict it as ``other class''. Thus, the new generated sample could belong to unknown classes.

\subsection{Others}
Apart from the above mentioned directions in representation learning across domains, which aim to improve the robustness in representation learning in real-world application scenarios. We also provide another two new directions: cross-domain few-shot learning and federated learning across domain.

The Cross-Domain Few-Shot Learning (CDFSL)~\citep{ICLR19-FSL,ICLR20-CDFSL,ECCV20-CDFSL,IJCAI21-CDFSL,ICCV21-CDFSL,ICLR22-CDFSL,ICLR22-CDFSL1,CVPR22-CDFSL} aim to understand and address the domain shift problem for few-shot learnign~\citep{ACMCS20-FSL-Survey,arXiv22-FSL-Survey}. Early, Chen et al.~\citep{ICLR19-FSL} found that existing few-shot learning methods fail to address the domain shift problem. Guo et al.~\citep{ECCV20-CDFSL} proposed the Broader Study of Cross-Domain Few-Shot Learning (BSCD-FSL) benchmark, which including image date from diverse domains, and evaluated plenty methods, e.g., meta-learning methods, transfer learning methods, few-shot learning method, and newly CDFSL methods. Tseng et al.~\citep{ICLR20-CDFSL} utlized feature-wise transformation layers to generate various feature distribution to improve the generalization capability of the learned model. Inspired by adversarial data augmentation~\citep{NeurIPS18-SDG-ADA}, Wang et al.~\citep{IJCAI21-CDFSL} proposed an adversarial meta task augmentation based framework to alleviate inductive bias problem and improve the robustness of the model under domain shift. Liang et al.~\citep{ICCV21-CDFSL} proposed a Novel-enhanced Supervised Autoencoder (NSAE) to help model capture broader variations of the feature representation and increase the discrimination and generalization capability of learned model on target domain. Recently, Das eta al.~\citep{ICLR22-CDFSL1} used contrastive learning to learn a generalizable representation and train a masking modeule to select features that can benefit target domain classification.

The Federated Learning Across Domain (FLAD)~\citep{peng2019federated,zhang2021federated,liu2021feddg,yao2022federated} aims to address the domain adaptation and domain generalization problems under the data privacy constraint. Although federated learning~\citep{zhang2021survey} have addressed data privacy and efficiency problem under the networks of distributed devices for deep learning. The learned model still suffer from domain shift problem which would degrade the robustness of the model in real-world application scenarios. In the context of federated learning, addressing domain adaptation challenges, Pent~\citep{peng2019federated} introduced a novel concept known as Unsupervised Federated Domain Adaptation (UFDA). This approach involves transferring knowledge from decentralized nodes to a new node with a distinct data domain, referred to as the target domain. To tackle domain shift in federated learning, they introduced Federated Adversarial Domain Adaptation (FADA), which employs a dynamic attention model to effectively manage the domain shift problem. Different from ~\citep{peng2019federated}, Zhang et al.~\citep{zhang2021federated} addressed the domain generalization problem under federated learning. They proposed FedADG framework, which including Adversarial Learning Network (ALN) for each node to align each source domain to the generated reference distribution, and a distribution generator in server to generated reference distribution.

\section{Privacy in Representation Learning Across Domains}
\label{privacy}
Deep neural networks are experiencing a surge in popularity across diverse domains including natural language processing, computer vision, rapid recognition, and recommendation systems.
The wide use of machine learning models, at the same time, also brings privacy concerns.
In recent years, customers are gradually provided with access to software interfaces (e.g., from Google, Microsoft, Amazon) that allow them to easily use machine learning tasks for their own applications.
However, if the training data are able to be recovered by some malicious users, serious issues can be caused by such information leaking, especially when the training data is sensitive, say, medical records.
Likewise, the parameters of the model, which are considered as proprietary information, should also be protected from being stolen with access to the interface.
In representation learning across different domains, due the information exchange among multiple domains, the risk of potential information leakage can be increased and requires more attention. 
Within this section, we initially provide a concise overview of common privacy challenges. Subsequently, our focus shifts towards addressing privacy concerns when undertaking cross-domain learning. Concluding, we shed light on less-explored challenges and potential avenues for forthcoming research.

\subsection{Privacy in Deep Learning}  

In this section, we briefly discuss what privacy threats exist for a deep learning model\cite{de2020overview} and roughly categorize them into three classes.

\subsubsection{Inference of the population}
\label{dai}
\begin{enumerate}
    \item \textit{Statistical disclosure} : given the trained model, the adversary tries to learn about the input of the model. Theoretically, to control statistical disclosure, it is required that applying the trained model should not reveal more knowledge about its training input than it is not applied.
    \item \textit{Model inversion} : The adversary seeks to deduce the sensitive attributes of the input to the model using the model's output, provided the input.
    \item \textit{Inference of class features} : Relaxing the sensitive attributes in model inversion to arbitrary class attributes which characters a particular task, the adversary tries to learn about the characters of each class with access to the model.
\end{enumerate}

\subsubsection{Inference of the training dataset}

This aspect focuses on the privacy of the individuals who provide data for the training of the model. 
\begin{enumerate}
    \item \textit{Membership inference} : given a model and an instance, the adversary discriminates if the instance was in its training dataset.
    \item \textit{Property inference} : In cases where training data are distributed across different clients, the adversary tries to infer characters that are true of a subset of the training data.
\end{enumerate}

\subsubsection{Inference of Model Parameters}

Different from previous aspects which focus on the data, this aspect concentrates on the privacy of the model itself. 
\begin{enumerate}
    \item \textit{Model Extraction} : The adversary tries to learn an equivalent or near-equivalent model with the access to the model (e.g., query access).
\end{enumerate}

\begin{figure}
    \centering
    \includegraphics[scale=0.50]{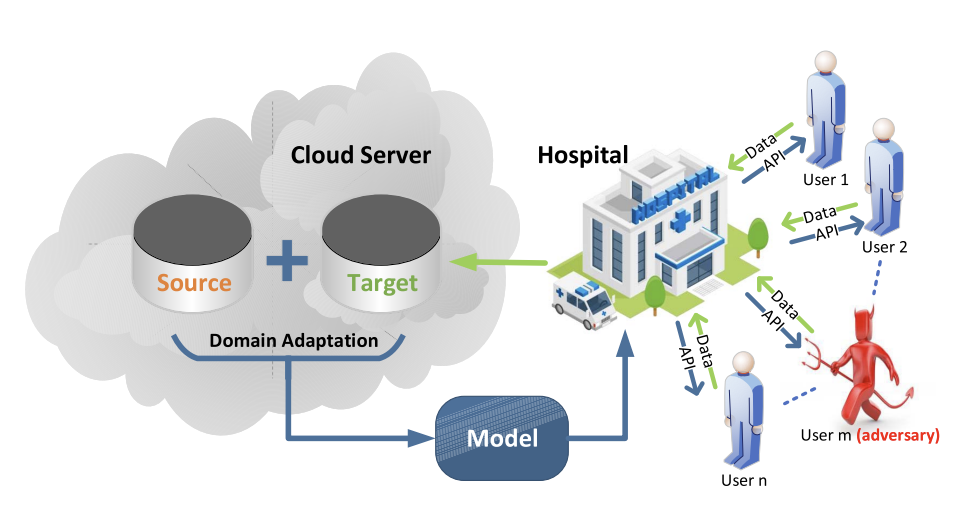}
    \caption{ An illustration of domain adaptation and the risk of privacy leakages. (image from \cite{wang2020deep})}
    \label{survey1}
\end{figure}

\subsection{Privacy issue in learning across different domains}
\textbf{Transfer learning} aims to enhance the learning of weakly labeled or unlabeled target domain with the acquired knowledge of source domain. 
Nowadays, the growing privacy protection awareness, which restricts access to different data sources makes traditional transfer learning more difficult to work.
Thus transfer learning technologies under strict privacy constraints should be developed to keep in step with changing regulations. 


Existing research efforts in the realm of privacy within transfer learning predominantly concentrate on safeguarding data privacy. Particularly, the concept of differential privacy, which offers a mathematically rigorous framework for data privacy, is commonly employed as the guiding principle.
Given a randomized algorithm $M$ with an output space $\Omega$ and a well-defined probability distribution, it is considered $(\epsilon, \delta)$-differentially private if, for all pairs of adjacent datasets $\boldsymbol{D}$ and $\boldsymbol{D}^{\prime}$ differing by a single data point $d$, and for all measurable sets $\omega \in \Omega$, the following inequality holds:
$$
\operatorname{Pr}[\boldsymbol{M}(\boldsymbol{D}) \in \omega] \leq \exp (\epsilon) \operatorname{Pr}\left[\boldsymbol{M}\left(\boldsymbol{D}^{\prime}\right) \in \omega\right]+\delta
$$ 

\noindent Differential privacy necessitates that upon obtaining the output of a differentially private algorithm $\textbf{M}$ applied to the confidential dataset $\textbf{D}$, there should be no means of ascertaining whether a specific data point $d$ is present in the private dataset $\textbf{D}$ or not. The parameter $\epsilon$ represents the privacy budget, signifying the extent of privacy inherent to the model. A diminished value corresponds to an augmented level of privacy.

Wu et al. (2022) and Wang et al. (2018) address the domain of differential privacy within multi-source transfer learning, encompassing the presence of multiple source domains. While prevailing endeavors in data privacy predominantly exploit differential privacy mechanisms within single-domain settings, these approaches often overlook the shifts in distribution across diverse domains.
To counter this limitation, Wang et al. (2018) introduce a methodology for differentially private multi-source hypothesis transfer learning. This technique assesses the interplay between source and target domains by utilizing hypotheses trained on the source domain. This strategy amplifies the refinement of target hypothesis acquisition by utilizing an accessible public unlabeled source dataset.
Concurrently, Wu et al. (2022) present the MultiSTLP algorithm, which engages in multi-source selection transfer learning with an emphasis on privacy preservation. Distinct from Wang et al. (2018), this approach leverages samples from a representative dataset culled from various sources, alongside unlabeled group probability samples within the target domain.

Domain adaptation, a subset of transfer learning, endeavors to adjust a model trained in a source domain to a target domain. This adaptation occurs when the distributions of the source and target domains differ yet maintain a certain level of interrelation. 
In scenarios where data from the source domain possess labels while data from the target domain remain unlabeled, the term "unsupervised domain adaptation" (UDA) is commonly used. Contemporary UDA techniques proficiently convey knowledge from the labeled source domain to the unlabeled target domain. However, in practical contexts, as exemplified in Figure 1, where disparate institutions furnish data for the source and target domains, the potential for privacy risks surfaces due to the information exchange between these distinct domains.

While most existing works fail to take this aspect into consideration, a few recent works \cite{wang2020deep, jin2021differentially, zhuang2021towards, an2022privacy, stan2021privacy} study how to perform unsupervised domain adaptation with privacy guarantees for training data. Wang et al. \cite{wang2020deep} propose to perform domain adaption with privacy assurance in an adversarial learning way and embed the differential private design into specific layers and learning processes. Different from  Wang et al. \cite{wang2020deep}, Jin et al. \cite{jin2021differentially} proposes a private correlation alignment approach for domain adaptation (PRIMA). 
In pursuit of privacy assurances, perturbations are introduced into the authentic covariance matrix, resulting in a confidential covariance matrix for the target domain. Moreover, during gradient-based model training, perturbations are applied to gradients to achieve a privacy-preserving model. Supplementary enhancements are implemented to address the constraints arising from the integration of these perturbation techniques.

Common unsupervised domain adaptation algorithms typically involve access to source domain data. However, in cases where heightened data privacy is imperative—such as with sensitive data like medical records—sharing source data with the target domain becomes unfeasible. Notably, Stan et al. \cite{stan2021privacy} and An et al. \cite{an2022privacy} delve into the UDA challenge under such privacy constraints.
To address this issue, Stan et al. \cite{stan2021privacy} employ an innovative approach where they encode information from source data into a prototypical distribution. This intermediate distribution serves as a means to align the distributions of the source and target domains. On a similar note, An et al. \cite{an2022privacy} propose an alternative strategy, where the feature distribution of the source domain, rather than the raw data, is shared. To counter membership inference attacks, An et al. \cite{an2022privacy} model the source feature distribution using Gaussian Mixture Models within a differentially private framework. This modeled distribution is then transmitted to the target domain for adaptation. Subsequently, the target client resamples differentially private source features from the Gaussian Mixture Models and adapts using state-of-the-art unsupervised domain adaptation techniques tailored for the target domain data.

Federated Learning adheres to data privacy constraints, typically disallowing data sharing between the server model and client models.
A line of research \cite{peng2019federated,sun2021partialfed,liu2021feddg,reddi2020adaptive,zhang2021federated} study domain adaptation or domain generalization under the federated learning setting, where only parameter transmission is allowed.
Different from traditional federated learning, it is more challenging because datasets may differ in both the sample and feature space. In such a setting, transferring the knowledge safely without violating the privacy constraints is required.
They are subject to the privacy constraint in federated learning setting, however, prohibiting data sharing is merely a basic approach to protect data privacy. For higher levels of data privacy protection, several recent works especially \cite{zhang2020privacy, gao2019privacy, song2020privacy, zhuang2021towards} focus on the privacy protection in such setting.
Different strategies are proposed for better adaptation under privacy constraints.
To protect client data privacy, homomorphic encryption based approaches \cite{zhang2020privacy, gao2019privacy, song2020privacy}  are proposed for effective knowledge transfer. 
More specifically, Zhang et al. \cite{zhang2020privacy} improves the commonly used maximum mean discrepancy with secure-aware technologies based on homomorphic encryption for effective knowledge transfer without compromising privacy. Song et al. \cite{song2020privacy} exploit homomorphic encryption to encrypt intermediate results which need transmitting to protect privacy. Gao et al. \cite{gao2019privacy} builds a federated transfer learning framework with an privacy preserving learning approach with two variants based on homomorphic encryption and secret sharing techniques. Zhuang et al. Zhuang et al. \cite{zhuang2021towards} further explore unsupervised domain adaptation within a federated learning context, with a primary emphasis on enhancing model performance through iterative aggregation of knowledge from the source domain via federated learning. In this approach, data privacy is safeguarded by exclusively transferring models, rather than raw data, between domains, omitting additional privacy-preserving techniques.
Different from other existing works, Kohen et al. \cite{pmlr-v162-kohen22a} first presents a general framework for transfer learning in the hybrid-model.

\section{Fairness in Representation Learning Across Domains}
\label{fairness}
Fairness in machine learning attracts a lot of attention nowadays. It protects interests of any individuals and groups in decision making or recommendation systems. In the context of across domains, fairness becomes more realistic but more difficult to achieve. In this section, we firstly talk about the unfairness circle and sources in across domain areas. We mainly focus on across domain in train/test situation, because across domain in training stage only, like class incremental continual learning, is very similar to traditional fairness settings. Then we introduce some approaches, bounding based, training based, causal inference based, to achieve fairness. At last, we discuss the current challenges and possible future directions.
\subsection{Unfairness Circle}
For transitional unfairness issues, the original bias emerges from imbalanced data, algorithms and user interactions. It is almost invariant in cross domain area. For data perspective, imbalanced data in source domain first introduce unfairness. For example, the dominant attribute classes would play more important role in prediction while ignore minor classes. In addition, though it satisfies great in source domain, it may cause serious unfairness when deploying in real world due to distribution shift. Algorithms, however, in order to fit the source domain distribution, it will inevitably skew the distribution in target domain and make less accurate and less fair predictions~\citep{chen2022fairness,coston2019fair,rezaei2021robust}. Moreover, the prediction result would further influence user preferences. The behavioural biases will aggregate the unfairness in data.

\subsection{Source of Unfairness in across domain area}
\begin{figure}
    \centering
    \includegraphics[width=0.8\textwidth]{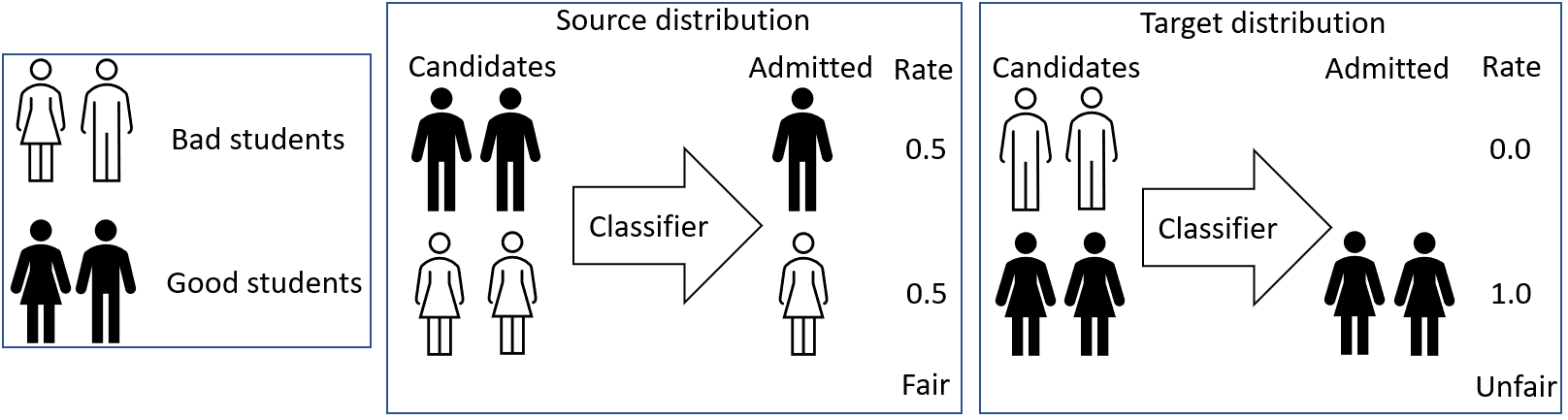}
    \caption{an example for \textbf{unfairness in covarite shift}. It shows the same prediction model for admission in different data distribution environments. In source domain, in order to satisfy same admission rate between males and females, the classifier will give the higher weight on gender attributes. However, in target domain, the model preference would cause the unfairness that female admission rate is higher than male's.}
    \label{fig:covarite unfairness}
\end{figure}

Unfairness issues in cross domain areas are mainly result from distribution shift~\citep{amodei2016concrete} or domain shift~\citep{sun2016return}. Distribution shift means that data distribution between training dataset and real world situation is different. In general, distribution shift can be classified as two types, including \textbf{covariate shift}~\citep{shimodaira2000improving, sugiyama2007covariate, gretton2009covariate} and \textbf{label shift}~\citep{scholkopf2012causal,pmlr-v80-lipton18a,azizzadenesheli2019regularized}. Covariate shift means the disparity of input variables distribution between different datasets. It will raise unfairness issues. For example, Fig.~\ref{fig:covarite unfairness} shows a classifier is fair in source distribution while it would produce unfairness results in target distribution since it would give higher weights on female attributes. As for the label shift, it means that feature distribution remains the same, the label distribution might changes. For example, one school may admit students with grades of B or higher, but another school will only admit students with grades of A. It is obvious that a fair model training on one school dataset will not be fair on the other school. However, a lot of works~\citep{chen2022fairness,coston2019fair,rezaei2021robust,oneto2020learning,singh2021fairness} try to mitigate such unfairness and improve the fairness transferability.

\subsection{Main work in fairness}
With different training stages, fairness can be divided into pre-process method, in-process method and post-process method. We summarize the related papers in table~\ref{tab:summary}. Pre-processing methods mitigate unfairness by modifying training data, such as data augmentation. For example, wang et al.~\citep{wang2021towards} tried to deal with fair domain adaptation classification problem. They generate supplementary samples in the source domain for few-shot classes to improve classification accuracy for minor class. However, in across domain area, there are not much work in pre-processing category since it is intractable to include all possible data distribution during training process by modifying data.
In-processing methods focuses on model itself to find a balance between accuracy and fair. post-processing approaches apply modification on model output while keep model fixed. Pre-processing methods mitigate unfairness by modifying training data, such as data augmentation. Most of existing work are in-processing methods because only in-processing approaches can directly optimize model to achieve fairness. In this section, we will mainly talk about in-processing methods. 

In processing approaches mainly consist of this three categories, \textbf{ bounding based, training based, causal inference based}.

\begin{figure}
    \centering
    \includegraphics[width=0.5\textwidth]{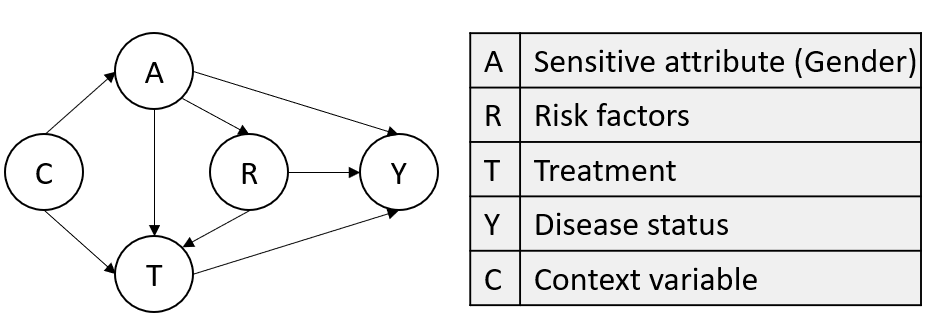}
    \caption{\textbf{Causal graph} for a disease treatment example. $A,R,T,Y$ are system variables. $C$ is the context variable. Arrow indicates the causality. For example, $T\to Y$ indicates that treatment will influence disease status.}
    \label{fig:causal graph}
\end{figure}

\subsubsection{Bounding based Approaches} The most popular and intuitive methods are bounding based approaches. They set up a fairness bound, e.g., fairness definition, to regulate representations or downstream tasks. Chen et al.~\citep{chen2022fairness} formulates a general upper bound for transferred group fairness violation subject to distribution shift, including covariate shift and label shift; Coston et al.~\citep{coston2019fair} made the problem more challenge. It focuses on a situation that sensitive attributes are only available either in source domain or target domain. They use demographic parity as their fairness metric. They use weighting methods to minimize empirical risk in order to bring group-specific prevalence closer together; zhao et al. ~\citep{zhao2021fairness} applied fairness in online meta-learning. It applies decision boundary covariance to regulate model.

\subsubsection{Causal inference based Approaches} Causal inference methods leverage the causal information to deal with the unfairness issues, especially in counterfactual fairness.
We can draw a common causal graph to illustrate the problem and fairness objective.
As shown in Figure.\ref{fig:causal graph}, $A,R,T,Y$ are system variables. $A$ is the sensitive attribute and $Y$ is the label for prediction. $C$ is the context variable, which indicates the domain distribution. In other words, $P(A,R,T,Y|C=0)$ and $P(A,R,T,Y|C=1)$ represent joint distribution as source and target domains, respectively. Now the fairness problem can be formulated as a minimization problem, i.e.
\begin{equation}
    f^*_{target} = arg\,min_{f\in F}\{(P(f(S) \neq Y|C=1) + G(f,P(Y,S|C=1)))\leq \epsilon)\}
\end{equation}

, where $S = {A,R,T}$ is the input features, and $G(f,P_{target})$ is the fairness constraint, and $f$ is the learnable function and $\epsilon$ is the hyperparameter controlling the intensity of constraints. In this equation, we have two terms. $P(f(S) \neq Y|C=1)$ describes the classification error. Minimizing it would result in higher accuracy. $G(f,P(Y,S|C=1))$ is the fairness constraint in target domain. Lower value indicates higher fairness achievement. For example, Demographic parity can be formulated as $$ |P(f(S)|A=0) - P(f(S)|A=1)| \leq \epsilon_0
$$
Singh et al.~\citep{singh2021fairness} studied the problem that the target distribution is unseen. They leverage the causal information to model data shifts and exploit conditional independencies in the data to estimate model accuracy and fairness metrics. They not only use group-fairness metrics, i.e. demographic parity and Equalized odds, but also extend to counterfactual fairness. Zhao et al.~\citep{zhao2020fair} inferred decision boundary covariance from fair causality and casual effect. They ensure fairness by controlling the covariance between the protected variables and the signed distance from the feature vectors to the decision boundary. 

\subsubsection{Training based Approaches}Training based methods aim to construct a training strategy to achieve fairness. For example, adversarial training strategy trains a fair representation to fool a sensitive discriminator. If the sensitive discriminator can not predict sensitive information, the representation and its downstream task can achieve fairness. Rezaei et al.~\citep{rezaei2021robust} converted fair prediction problem to a competition between an adversary to fairly minimize predictive loss and an adversary to maximize that objective loss. Rezaei et al.~\citep{rezaei2021robust} proposed a approach to obtain a predictor that is robust to covariate shift and satisfy target fairness requirements. They formulate a penalty term that can evaluate the fairness in target distribution and optimize the problem based on that. While Mandal et al.~\citep{mandal2020ensuring} moved a step forward. They train a fair classifier that is robust to any distributions that could represent any arbitrary weighted mixtures of training distributions. They formulate a min-max objective function. On the one hand, it minimizes distributionally robust training loss; on the other hand, it finds a fair classifier with respect to a class of distribution. Hong et al.~\citep{Hong21Federated} considered federated learning situation. The debiasing process does not involve accessing users' sensitive group information, ensuring their privacy. Additionally, users have the freedom to choose whether to participate in the adversarial component, particularly when they have concerns about privacy or computational costs.

Multi-task training strategy can also formulate fair and robust representations. Oneto~\citep{oneto2020learning} focused on generate robust and fair representations. they learn fair and transferable representations by leveraging multi-tasks learning with guarantees on generalization to the target domain.

\begin{table*}[t]
\centering
\small
\caption{Summary of the typical fairness in across domains scenarios. Goal refers to the target task. Specifically, "Data" means making training data unbias; Bound means setting up a fairness upper bound; prediction is predicting labels; and representation refers generate fair latent embeddings. In the fairness metric column, DP denotes demographic parity~\citep{dwork2012fairness} or statistic parity and EO refers to Equal oppotunity~\citep{hardt2016equality} and CF means counterfactual fairness~\citep{kusner2017counterfactual}, and DBC is Decision Boundary Covariance~\citep{zafar2017DBC}. }
\begin{tabular}{c|c|c|c|c}
\toprule
Methods                     & Goal  &{Fairness metric}  & Phase &{Condition} \\
\midrule 
\cite{wang2021towards}      & Data  &   -            & Pre   & few-shot DA \\
\cite{chen2022fairness}     & Bound &   DP, EO      & In    & -\\
\cite{coston2019fair}       & Prediction & DP & In & Missing sensitive attributes          \\
\cite{zhao2021fairness}     & Prediction & DP,EO & In& Online learning\\
\cite{singh2021fairness}    & Prediction & DP, EO, CF   & In & Unseen test set\\
\cite{zhao2020fair}         & Prediction & DBC & In & few shot\\
\cite{rezaei2021robust}     & Prediction & EO & In & target data with unknown label\\
\cite{mandal2020ensuring}   & Prediction & DP, EO & In & - \\
\cite{Hong21Federated}      & Prediction & EO& In& federated learning\\
\cite{oneto2020learning}    & Representation & DP & In & -\\


\bottomrule
\end{tabular}
\label{tab:summary}
\end{table*}

%

\section{Explainability in Representation Learning Across Domains}
\label{explainability}
Interpretable and explainable machine learning models emerge from a situation that human is model agnostic. In order to understand why model works well or not, interpretable and explainable models are proposed to help explain machine learning models in an intuitional way. Explainability has been studied in many areas. It can be defined as "the ability to explain or to present in understandable terms to a human"~\citep{doshi2017towards}. In the context of representation learning across domains,
we hope we can know what a model learn from source domain and how a model apply it to target domain. Thus, when a model is deployed in real world that work poor or violate some safety~\citep{ICII17-AutoDriving} or ethnicity~\citep{news10-face} rules, we can easily troubleshoot them.

In this section, we first introduce the goal of explainability models and some concepts, and then we summarise current work in explainable and interpretable work in across domain area. We divide them into explainable models and interpretable models. At last, we will talk about its future directions. 

\subsection{Concepts}
\subsubsection{interpretbility}
Interpretability requires models to give accurate and meaningful outputs, i.e., it not only answers what the results are but also explains why the results are presented. Specifically, interpretable models reflect physical constraints, monotonicity, additivity, causality, sparsity and/or other desirable properties~\citep{rudin2019stop,carvalho2019machine,marcinkevivcs2020interpretability}.
In the context of representation in across domains, the interpretability of models leverage intrinsically interpretable designs to explain predictions or generations.
\subsubsection{explainability}
explainability refers to explain an existing black-box model post hoc. Explainable model is independent to target model that we try to explain. Bascially, it aims to analyze the results that a model gives in a textual, visual or symbolic way~\citep{lipton2018mythos}. It can be model agnostic or model specific. In either way, it should satisfy Faithfulness, Consistency, Comprehensibility, Certainty, Representativeness.~\citep{carvalho2019machine}

\subsection{Main work}

\subsubsection{Interpretable model}
\begin{figure}
    \centering
    \includegraphics[width=0.6\textwidth]{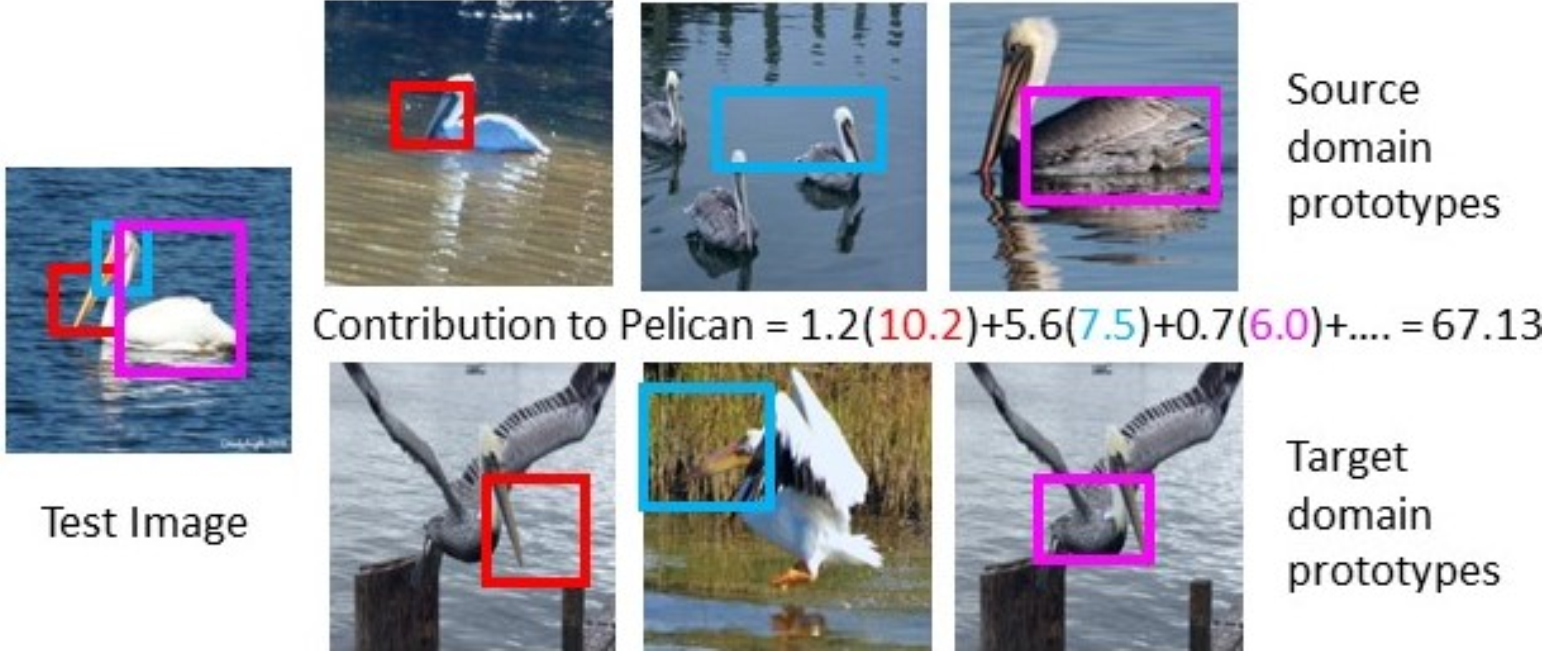}
    \caption{XSDA-Net instance-level explanation. It extract some image prototypes from source domain and target domain. According to comparing test images with these prototypes, the similarity scores are forwarded to prediction. Image from~\citep{kamakshi2022explainable}.}
    \label{fig:kamakshi2022explainable}
\end{figure}
Interpretable models can be divided into instance-level, group-level and class-level models in terms of interpretable objects. Instance-level models can give explanations for each samples.
Kamakshi et al.~\citep{kamakshi2022explainable} proposed XSDA-Net to explain the prediction of every test instance in terms of prototypes in the source and target samples. XSDA-Net extracts prototypes and calculates similarity with these prototypes, like Fig.~\ref{fig:kamakshi2022explainable}. Zunino et al.~\citep{CVPRW21-Exp-DG} explains their across domain classification network by means of a saliency map conveying how much each pixel contributed to the network prediction. Raab et al.~\citep{raab2022domain} also adapted saliency map to interpretable their model. Though saliency map is wildly used in interpretable models, it has some limitations. For example, saliency map is unable to give sufficient explanations for those false predictions since it focuses on the same pixels when it predicts correctly. 

Other across domain generating task also involves interpretability, such as semantic autoencoder (SAE), Gernerative adversarial network(GAN). Back from~\citep{kodirov2017semantic}, Kodirov et al. tried to align latend feature space with semantic feature space. In this way, image features can be translated into semantic level, vice versa. Conditional GAN~\citep{mirza2014conditional} also wants to generate images while controlling some semantic information. Gao et al.~\citep{gao2020Zero} used variational autoencoder (VAE) and generative adversarial network (GAN) jointly to generate unseen features. Li et al.~\citep{Li_2019_Leveraging} proposed LisGAN, which can directly generate the unseen features from random noises which are conditioned by the semantic descriptions. These generative methods inherit explainability.

\subsubsection{explainable model}
\begin{figure}
    \centering
    \includegraphics[width=0.4\textwidth]{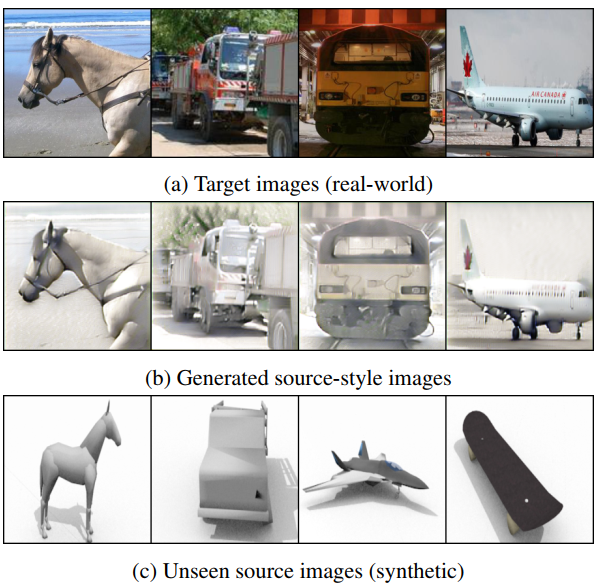}
    \caption{Visualization of adapted knowledge in unsupervised domain adaptation on the VisDA dataset~\citep{peng2017visda}. Fig.a shows the target images. Fig.b are images generated from fig.a. Fig.c is the source images which is inaccessible during generating process. Image from~\citep{hou2021visualizing}.}
    \label{fig:hou2021visualizing}
\end{figure}

Explainable models aim to explain black box prediction models. It can be divided as model-agnostic and model-specific categories. However, this topic in across domain is not studied very much. Hou et al.~\citep{hou2021visualizing} is model-agnostic one, which visualizes the adapted knowledge difference with image translation. Like Fig.~\ref{fig:hou2021visualizing}, they portray the knowledge learned from different UDA models and find that these generated images are able to capture the difference between source and target domains.

\section{Future Research Directions}\label{future}
We have introduced our trustworthy in representation learning across domains framework in Sec.~\ref{Concepts} and go through the existing methodologies developed for each concepts in Sec.\ref{robustness}, \ref{privacy}, \ref{fairness}, \ref{explainability}. Instead of only reviewing the existing methods, we also try to o share some insights on future research directions in our trustworthy framework.

\subsection{Future Research Direction of Robustness}
We have reviewed several directions from different scenarios for representation learning across domains in terms of Robustness. In this section, we provide four potential research directions.

\textbf{Novel Categories Recognition.}  In representation learning across domains, a key task within the open-set framework is identifying target data that lies outside the source label space. These endeavors~\citep{ECCV18-OSDA,ICCV19-OSDA,ECCV2020-DA-UniDA,CVPR21-UniDA,CVPR21-DG-OSDG,ICLR22-DG-OSDG} typically center on mining valuable insights from data to aid in classifying target data into either known or unknown categories. In some cases, augmenting data with more information can facilitate the development of robust representations, pertinent to real-world applications. For instance, Jiang et al.~\citep{ICCV21-DA-OSDA} introduced semantic attribute annotations for the source domain. This approach yielded a more resilient model compared to existing methods. Another avenue for inducing meaningful information involves leveraging vision and language models~\citep{jia2021scaling,li2021align,radford2021learning}. These models are trained using image-caption pairs, offering beneficial impacts on various computer vision tasks, including object detection~\citep{zhao2022exploiting} and object classification~\citep{xu2022generating}. This alternative direction has the potential to enrich the representations and enhance the performance of models in a range of applications.


\textbf{Few Labeled Data.} In the realm of representation learning across domains, a prevailing assumption is that target data remains unlabeled. However, fully annotating all target data can be both time-consuming and costly in practical applications. In many scenarios, having access to a small set of labeled target data is feasible and economical. Exploring effective strategies to maximize the utility of these limited labeled target data represents a promising avenue to enhance the robustness of learning representations. This direction holds potential for further advancements in representation learning across domains.

\textbf{Feature Generation.} The efficacy of data augmentation in bolstering the robustness of representation learning across diverse domains has been firmly established~\citep{NeurIPS20-SDG-MEADA,yue2019domain,AAAI20-DG-DA,CVPR21-DG-OSDG}. These methods often focus on generating new data to enrich the diversity of the original data distribution. However, the concept of feature generation has received relatively limited attention. Only a handful of studies have explored this aspect, yielding substantial improvements. For instance, Xu and collaborators~\citep{xu2022generating} introduced a model based on conditional variational autoencoders to generate new features that complement the original feature distribution. Li et al.~\citep{li2020adversarial}, on the other hand, harnessed conditional Wasserstein generative adversarial networks to produce a variety of distinctive and discerning features. These efforts emphasize the untapped potential of enhancing features through data augmentation, offering a unique perspective on advancing representation learning across domains.

\textbf{Privacy Issue based Cross-Domain Tasks.} When People and society are paying more and more attention to data privacy, federated learning is proposed to address this issue. But this also leads to the learning model is vulnerable in real-world applications. To improve the robustness of the model under the requirement of protecting data privacy, several attempts have been done. For example, solving the unsupervised domain adaptation~\citep{peng2019federated} and domain generalization~\citep{zhang2021federated} via federated learning framework. There still existing plenty of cross-domain tasks, e.g., various tasks in domain adaptation and domain generalization, to be solved through federated learning framework.  

\textbf{Foundation Models}. Recently, foundation models such as Contrastive Language-Image Pre-Training (CLIP)\cite{radford2021learning} and Segment Anything Model (SAM)\cite{kirillov2023segment} have demonstrated remarkable capabilities in generalization, particularly in zero-shot and few-shot learning scenarios. However, these foundation models still face challenges in cross-domain learning tasks. Several studies~\citep{ge2022domain,zara2023autolabel,shu2023clipood} have highlighted that the generalization of these foundation models needs improvement, especially when dealing with tasks involving significant domain shifts between the source and target domains. Consequently, a promising avenue of research involves enhancing the generalization of foundation models for downstream tasks in the presence of domain shifts. This represents an emerging direction worth exploring.

\subsection{Future Research Direction of Privacy}
While most existing works pay attention to the data privacy on unsupervised domain adaptation, the data privacy issue in other existing domain adaption and domain generalization tasks mentioned in Sec.~\ref{background} also deserve attention such as: semi-supervised domain adaption, open-set domain adaption, and open-set domain generalization. Besides, in existing works, the focus of privacy in representation learning across domains is data privacy. However, other types of privacy, as discussed in Sec.~\ref{dai} are still under-explored in the context of learning across different domains. There are still many challenging problems remaining to be explored in the future.

\subsection{Future Research Direction of Fairness}
\textbf{Fairness Definition and Evaluation.} The fairness in across domain area is still a emerging topic. The definition of fairness mostly rely on traditional metrics, i.e. demographic parity~\citep{dwork2012fairness}, equal odds~\citep{hardt2016equality}. However, most work currently focus on a fixed target domain. In realistic, target domain may varied and even changes every time. Thus, how to define a general fairness to cover the online situation. In addition, in some tasks, such as face recognition, it is important to achieve fairness in both source and target domains. How to define such fairness which can cover both domains.

\textbf{Interpretable Fairness.} Fairness models are working on achieve fairness in mathematical or statistical way no matter in independent and identically distributed data or across domain data. How can we use some explainable way, such as causal graph, saliency map, to interpret models so that it can further direct human to make non-discriminatory judgments.

\textbf{Fairness in Foundation Models.} Foundation models such as the Segment Anything Model (SAM)~\citep{kirillov2023segment} and CLIP-based models exhibit remarkable zero-shot capabilities. However, their effectiveness is often hindered by inherent biases and imbalances present in their training data, leading to challenges in ensuring fairness across various contexts. Several recent studies~\citep{Seth2023CVPR, wolfe2023contrastive} have diligently quantified the extent of unfairness exhibited by these multi-modal models. Addressing these fairness concerns requires innovative research directions that focus on enhancing the fairness of foundation models for downstream tasks, even in the face of domain shifts. A promising avenue involves strategies to mitigate these issues. For instance, one approach is through prompt learning. This technique encompasses the design of carefully crafted prompts or the acquisition of dynamically adaptable soft prompts, enabling the foundation model to consciously reduce attention towards sensitive attributes. This adjustment can significantly contribute to improving the fairness performance of the model.

\subsection{Future Research Direction of Explainability}
\textbf{Explainable Metrics.} Explanability of existing models are unquantifiable. For example, these works~\citep{raab2022domain, CVPRW21-Exp-DG} use saliency map to explain their models. However, it is intractable to prove whether they give great explanations or not. Thus, a metric for explanability is needed to measure those explainable models.

\textbf{High-level explanations.} No matter interpretable models~\citep{kamakshi2022explainable,CVPRW21-Exp-DG} or explainable models~\citep{hou2021visualizing} explain across domain models in a low-level way, such as saliency map, Although it can bring some latent information for us to understand a model, it is still unable to troubleshoot misclassified samples. Therefore, high-level explanations, like nature language, semantic information, are necessary to better assist human to understand models.

\section{Conclusion}\label{conclusion}
In this survey, we instantiate the trustworthy AI core concepts to representation learning across domains. We introduce four foundational crucial concepts of trustworthy AI system, i.e., robustness, privacy, fairness, and explainability, to build our trustworthy representation learning across domains framework. For each concepts, we present a comprehensive overview of related methodologies to help reader understand, how each concepts is addressed. Additionally, we provide comprehensive details of advanced technologies to ensure a thorough understanding of the research. Furthermore, we delve into potential future research directions related to each concept, aiming to inspire further advancements in trustworthy representation learning across domains.


\bibliographystyle{unsrtnat}
\bibliography{sample-base}  






\end{document}